\documentclass[10pt,twocolumn,letterpaper]{article}

\makeatletter
\@namedef{ver@everyshi.sty}{}
\makeatother
\usepackage{array}
\usepackage{amsmath}
\usepackage{amssymb}
\usepackage{amsfonts}       
\usepackage{arydshln}
\usepackage{algorithm}
\usepackage{adjustbox}
\usepackage[noend]{algpseudocode}
\usepackage[accsupp]{axessibility}
\usepackage{booktabs}
\usepackage{caption}
\usepackage[pagenumbers]{cvpr}
\usepackage{epsfig}
\usepackage{graphicx}
\usepackage[pagebackref,breaklinks,colorlinks]{hyperref}
\usepackage[misc,geometry]{ifsym}
\usepackage{listings}
\usepackage{multirow}
\usepackage{multicol}
\usepackage{mathtools}
\usepackage{makecell}
\usepackage{pifont}
\usepackage{stmaryrd}
\usepackage{subfiles} 
\usepackage{svg}            
\usepackage{times}
\usepackage{url}            
\usepackage{nicefrac}       
\usepackage{microtype}      
\usepackage{xcolor}         
\usepackage{xr}
\usepackage[capitalize]{cleveref}

\newcolumntype{P}[1]{>{\centering\arraybackslash}p{#1}}
\newcommand{\LSI}[0]{SLI{}}
\newcommand{\SLI}[0]{SLI{}}

\newcommand{\keypoint}[1]{\vspace{0.12cm}\noindent\textbf{#1}\quad}
\newcommand{\cut}[1]{}

\crefname{section}{Sec.}{Secs.}
\Crefname{section}{Section}{Sections}
\Crefname{table}{Table}{Tables}
\crefname{table}{Tab.}{Tabs.}

\crefformat{footnote}{#2\footnotemark[#1]#3}

\def\cvprPaperID{689} 
\def\confName{CVPR}
\def\confYear{2023}

\begin{document}

\title{SketchXAI: A First Look at Explainability for Human Sketches}

\author{%
  Zhiyu Qu$^{1, 3}$\:\; Yulia Gryaditskaya$^{1}$\:\; Ke Li$^{1, 2}$\:\; Kaiyue Pang$^{1}$\:\; Tao Xiang$^{1, 3}$\:\;Yi-Zhe Song$^{1, 3}$\\[0.2cm]
  $^{1}$SketchX, CVSSP, University of Surrey
  $^{2}$Beijing University of Posts and Telecommunications \\
  $^{3}$iFlyTek-Surrey Joint Research Centre on Artificial Intelligence \\
  \small{\texttt{\{z.qu, y.gryaditskaya, kaiyue.pang, t.xiang, y.song\}@surrey.ac.uk}} \\
  \small{\texttt{\{like1990\}@bupt.edu.cn}}\\
}

\twocolumn[{%
\renewcommand\twocolumn[1][]{#1}%
\maketitle

\begin{center}
    \centering
    \vspace{-0.4cm}
    \includegraphics[width=\textwidth]{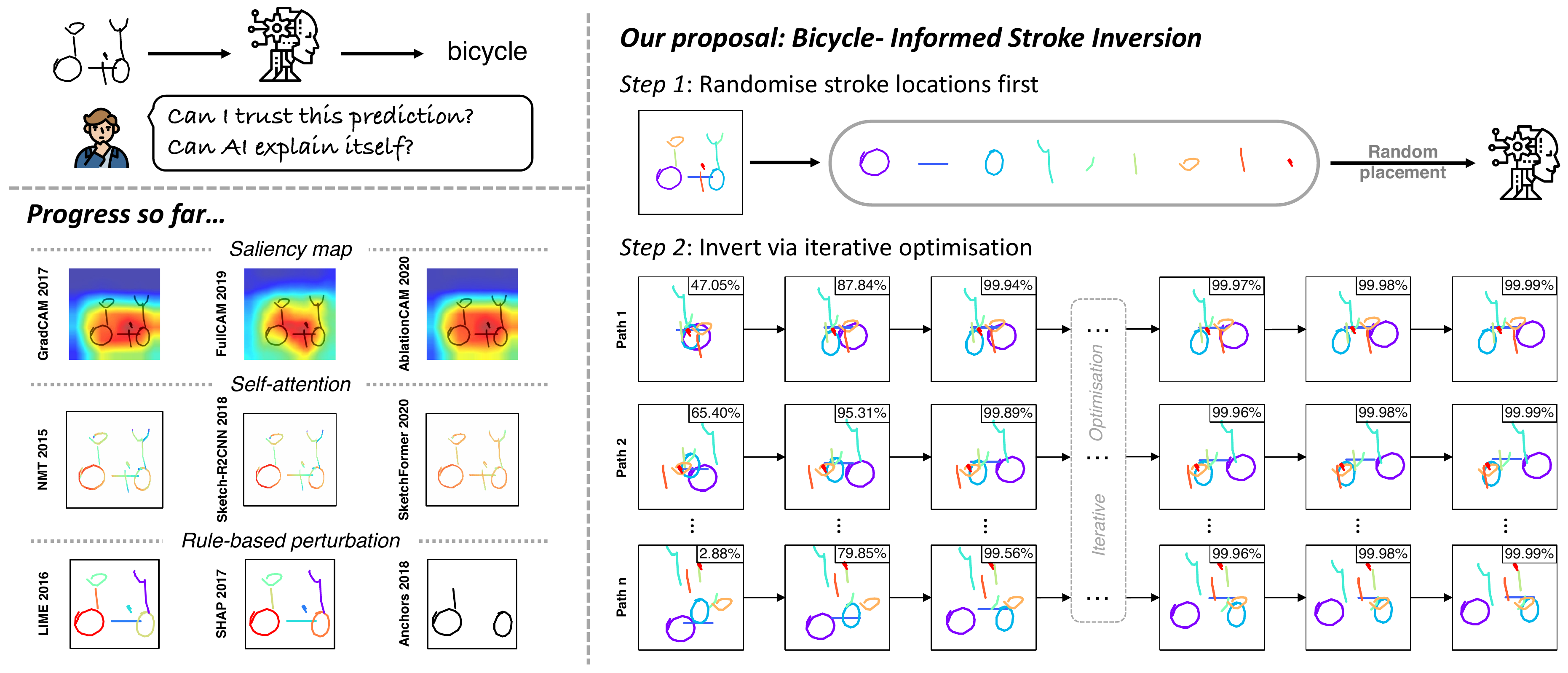}
    \captionof{figure}{\textbf{Explainability, but for human sketches.} We demonstrate a new methodology for explaining AI decisions on human sketch data. Instead of one static explanation per instance as in existing works, our proposed method supports generating infinitely many dynamic explanations with each showcasing how an AI classifier responds to a different view of the same sketch input. This is made possible by seeing stroke as the fundamental block to explain a sketch and randomising every stroke location to introduce stochasticity at every case. The infinite varieties of explanation paths allows humans to enjoy a wider coverage on how AI functions, and therefore to better scrutinise AI.}
    \label{fig:overview}
    \vspace{0.3cm}
\end{center}%
}]

\begin{abstract}
This paper, for the very first time, introduces human sketches to the landscape of XAI (Explainable Artificial Intelligence).
We argue that sketch as a ``human-centred'' data form, represents a natural interface to study explainability.
We focus on cultivating sketch-specific explainability designs.
This starts by identifying strokes as a unique building block that offers a degree of flexibility in object construction and manipulation impossible in photos.
Following this, we design a simple explainability-friendly sketch encoder that accommodates the intrinsic properties of strokes: shape, location, and order.
We then move on to define the first ever XAI task for sketch, that of stroke location inversion (\LSI). Just as we have heat maps for photos, and correlation matrices for text, \LSI{} offers an explainability angle to sketch in terms of asking a network how well it can recover stroke locations of an unseen sketch.
We offer qualitative results for readers to interpret as snapshots of the \LSI{} process in the paper, and as GIFs on the project page.
A minor but interesting note is that thanks to its sketch-specific design, our sketch encoder also yields the best sketch recognition accuracy to date while having the smallest number of parameters.
The code is available at \url{https://sketchxai.github.io}.
\end{abstract}

\section{Introduction} \label{sec:intro}
It is very encouraging to witness a recent shift in the vision and language communities towards Explainable AI (XAI) \cite{van2004explainable, ribeiro2016should, lipton2018mythos, zhou2016learning, bahdanau2014neural, arrieta2020explainable, tjoa2021survey}.
In a world where ``bag of visual words'' becomes ``bag of tricks'', it is critically important that we understand why and how AI is making the decisions, especially as they overtake humans on a series of tasks \cite{silver2016mastering, he2015imagenet, jumper2021highly, ouyang2022training}.

\vspace{1.5mm}

XAI research to date has focused on two modalities: photo \cite{escalante2018explainable, molnar2022interpretable, zhang2020explainable, linardatos2021explainable} and text \cite{shi-etal-2016-string, voita2018context, linzen2016assessing, liu2021explainaboard, galassi2020attention}.
Great strides have been made in the XAI for the photo domain, with the trend of going from heat/saliency maps \cite{selvaraju2016grad, chattopadhay2018grad, srinivas2019full, simonyan2013deep, zeiler2014visualizing} to the rules/semantics-oriented approaches \cite{kim2018interpretability, karras2020analyzing, shen2020interpreting}.
The text side is captivating due to the flexibility of sentence construction.
Early works in text models explainability also started with visualisations \cite{simonyan2013deep, zeiler2014visualizing, adebayo2018sanity}, moving onto linguistic phenomena \cite{linzen2016assessing, blevins2018deep, williams2018latent}, and most recently to attention \cite{strubell2018linguistically, hao2020self, rogers2020primer}.

\vspace{1.5mm}

In this paper, we make a first attempt at XAI for human freehand sketches.
The ``why'' we hope is obvious -- sketches are produced by \textit{humans} in the first place(!), from thousands of years ago in caves, and nowadays on phones and tablets.
They are uniquely expressive, not only depicting an object/scene but also conveying stories -- see a ``Hunter and Arrows'' here for a story dating back 25,000 years in France\footnote{\url{https://www.worldhistory.org/Lascaux_Cave/}}.
They, therefore, form an ideal basis for explainability which is also \textit{human-facing}.

\vspace{1.5mm}

The sketch domain is uniquely different from both of the well-studied photo and text domains.
Sketch differs from photo in that it can be freely manipulated, while photos are rigid and hard to manipulate.
This is largely thanks to the stroke-oriented nature of sketches -- jittering strokes might give the ``same'' sketch back, jittering pixels gives you a ``peculiar''-looking image.
Sketches have the same level of flexibility in semantic construction as text: strokes are the building block for a sketch as words are for text.
With these unique traits of sketch, the hope of this paper is to shed some light on what XAI might look for sketch data, and what it can offer as a result to the larger XAI community.
This, however, is only the very first stab, the greater hope is to stir up the community and motivate follow-up works in this new direction of ``human-centred'' data for XAI.

\vspace{1.5mm}

With that in mind, we focus our exploration on what makes sketches unique -- yes, \textit{strokes}.
They allow for flexible object construction and make sketches free to manipulate.
We then ask how strokes collectively form objects. For that, we identify three inherent properties associated with strokes: shape, location, and order.
These three variables define a particular sketch: \textit{shape} defines how each stroke looks like, \textit{location} defines where they reside, and \textit{order} encodes the temporal drawing sequence.

\vspace{1.5mm}

Our first contribution is a sketch encoder, that factors in all the  mentioned essential properties of strokes.
We hope that this encoder will build into its DNA how strokes (and in turn sketches) are represented, and therefore be more accommodating when it comes to different explainability tasks (now and in the future) -- and for this, we name it SketchXAINet (``X'' for E$\underline{X}$plainability).
We are acute to the fact that explainability takes simple forms \cite{MILLER20191}, so we refrained from designing a complicated network.
In fact, we did not go any further than introducing a branch to encode each of the three stroke properties (shape, location, and order), and simply feed these into a standard transformer architecture with a cross-entropy loss.
Interestingly, however, just with this simple architecture, we already observe state-of-the-art sketch recognition performance improving on all prior arts.

\vspace{1.5mm}

With an explainability-compatible sketch encoder in place, we now want to examine if we can actually make anything explainable.
First and foremost, of course, sketch explainability can be performed in the form of a heat map \cite{selvaraju2016grad, chattopadhay2018grad, srinivas2019full} -- just treat sketches as a raster image and we are done. This, however, would be entirely against our very hope of spelling out sketch-specific explainability -- the ``explainability'' one can obtain there is \textit{at best} at the level of photo heatmaps (see \cref{fig:overview}).

\vspace{1.5mm}

Instead, we utilise our sketch encoder and put forward the first XAI task for sketch -- that of stroke location inversion (\LSI) (see \cref{fig:overview,fig:recovery}). We study two types of tasks: recovery and transfer.
Intuitively, during the recovery, we ask our optimisation procedure to jitter the stroke locations to \emph{recover} sketch so that it belongs to the same class as the original sketch.
During the transfer task, we ask our optimisation procedure to jitter the stroke locations to obtain a sketch that belongs to a new class that we pass as input to the optimiser.
The idea is then that how well the network has learned is positively correlated with how well it does at this inversion task, and that explainability lies in visualising this process.
So, in addition to heat maps for photos, and correlation matrices for text, for sketch, we now have visualisations, that theoretically be manifested of infinite variety, and in the form of a video/GIF to capture the \LSI{} process.
We finish by playing with variants of the proposed SLI: (i) sketch recovery, to offer insights on category-level understanding of a learned encoder, \ie, reconstructing a sketch to the same category, and (ii) sketch transfer, to shed light on cross-category understanding, \ie, using strokes of one category to reconstruct another.

\vspace{1.5mm}

Our contributions are as follows: (i) we argue for sketches to be introduced to the field of XAI, (ii) we identify strokes as the basic building block and build a sketch encoder, named as SketchXAINet, that encapsulates all unique sketch properties, (iii) we introduce stroke location inversion (SLI) as a first XAI task for sketch, (iv) we offer qualitative results of the inversion process and deliver best sketch recognition performance as a by-product.

\section{Related work} \label{sec:related_work}

\keypoint{Raster and vector sketch encoders.} Sketch contains high-level human understanding and abstraction of visual signals and is a distinctive modality to photos.
Many of the previous works \cite{klare2010matching,schneider2014sketch, li2014fine,qi2015making, sangkloy2016sketchy, pang2019generalising, pang2020solving, liu2020scenesketcher, wang2021sketchembednet}, however, treat sketches with no difference to photos -- they operate on raster format and feed them into contemporary CNNs for visual learning.
Facilitated by the availability of sketch datasets with stroke-level information \cite{ha2017neural, ge2020creative}, there is an ongoing trend of works that turn to model sketch as a temporal sequence of vector coordinates, hoping to open up new research insights for downstream sketch tasks \cite{muhammad2018learning,usp, song2018learning, lopes2019learned,sketchbert,das2020beziersketch,yang2021sketchaa,li2022free2cad,yang2022finding}. Along with this representation change on sketch data is also the backbone upgrade, from CNN to Transformer \cite{sketchbert,sketchformer}, the choice of which we also embrace in constructing our proposed sketch encoder. Scarcely few existing works have anchored their focus on the explainability of sketch models, with \cite{muhammad2018learning} \cite{alaniz2022primitives} being moderately relevant to our best knowledge. At a high level, both works, just like ours, explore the impact of strokes on forming a sketch object. But instead of studying sketch abstraction, \ie, how strokes can be deleted or simplified without altering the holistic semantic meaning,  we leverage the free-hand stroke itself as a building block to understand sketch model explainability.

\keypoint{Ante-hoc and post-hoc explainability methods.}
Several recent surveys and books discuss explainability methods in detail \cite{molnar2022interpretable, antoniadi2021current,holzinger2019causability, arrieta2020explainable}.
Explainability methods are often split into two groups: \emph{ante-hoc} \cite{letham2015interpretable,caruana2015intelligible} and \emph{post-hoc} \cite{zeiler2014visualizing, zhou2016learning, ibrahim2019global, ribeiro2016should, lundberg2017unified, ribeiro2018anchors} methods.
Ante-hoc methods are inherently and intrinsically interpretable, while post-hoc methods require designing separate techniques to provide probes into model explainability.
The former, also known as the white/glass box approach, is preferable under the context of explainability, but limited by a few specific choices of instantiations, \eg, decision trees \cite{wan2021nbdt}, generalised additive models \cite{agarwal2021neural}.
The latter being less transparent has no restrictions on model learning and therefore often achieves better test-time task performance. Achieving the optimal trade-off of such is then the core to both schools of explainable AI \cite{holzinger2019causability, arrieta2020explainable}. Our proposed sketch explainability method SLI is post-hoc, but facilitated by a tailor-designed, less black-box (ante-hoc alike) sketch encoder (that allows reasoning over a stroke-based decision into shape, location, and order). Notably, our final sketch model achieves state-of-the-art recognition performance.

\keypoint{Counterfactual explanation and adversarial attack.} Our post-hoc explainability strategy SLI of ``relocating first, recovery later" is also reminiscent of a specific AI explainability genre -- counterfactual explanation (CE) \cite{wachter2017counterfactual, karimi2020survey, looveren2021interpretable}.
CE aims to identify what are the minimal input changes for a model to make a different visual decision. The model is then explained towards credibility if these changes are key in defining the
underlying visual concept. In this sense, SLI identifies the stroke location changes that matter (\eg, the tires and the front handle for a bicycle in \cref{fig:overview}) through multiple randomly initialised stroke inversion tasks (because important strokes gets highlighted across trials).
Closely related to CE is another field known of adversarial attack \cite{szegedy2013intriguing, goodfellow2014explaining, moosavi2017universal, baluja2017adversarial}, which aims at the generation of adversarial examples (AE) having \textit{imperceptible differences} to human vision but results in completely different AI predictions. Conceptual connections between CE and AE have been extensively discussed in the literature \cite{pawelczyk2021connections, browne2020semantics, wachter2017counterfactual}, where \cite{pawelczyk2021connections} suggests that AE is part of a broader class of examples represented by CE. Our proposed SLI also generates sketch samples that dictates a prediction change. We however model the generation process via the spatial reconfiguration of strokes, which is intrinsically distinctive to AE -- the movement of strokes is less likely to be imperceptible changes to human viewers compared with those by local pixel jittering.

\begin{figure*}
    \centering
    \includegraphics[width=\linewidth]{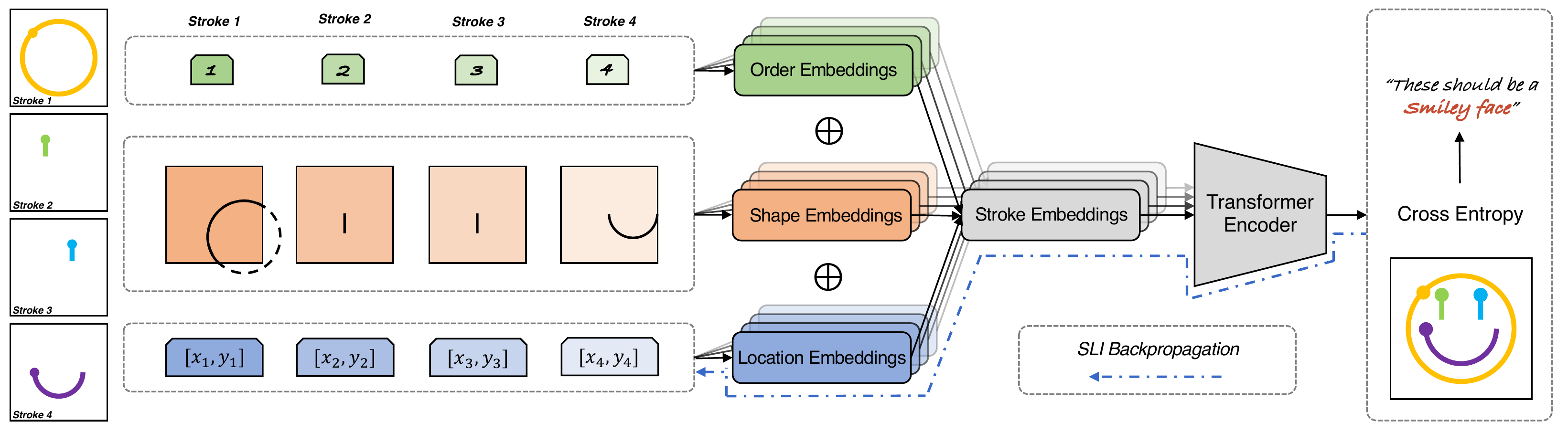}
    \caption{
    \textbf{SketchXAINet architecture.} We build a sketch classifier upon stroke vectors rather than raster pixels. All strokes are decomposed into three parts -- order, shape and location. We use a bidirectional LSTM, a linear model and a learnable time embedding matrix to encode such decomposed stroke representation respectively. The dashed line refers to the gradient flow of the location parameters when we generate explanations by \LSI{} with a trained classifier.
    }
    \label{fig:framework}
    \vspace{-0.2cm}
\end{figure*}

\section{Methodology} \label{sec:formulation}

In this section, we first introduce our classification model which is designed around strokes as sketch building blocks. We then introduce our method for model explainability.

As a pre-processing step, we simplify all sketches by the RDP algorithm \cite{douglas1973algorithms}.
For each stroke $s_i$ consisting of $k$ points, $\{s_{i, 1}, s_{i, 2}, ..., s_{i, k-1}, s_{i, k}  \}$, we identify three inherent properties in $s_i$ and learn respective descriptor for each: location $l_i$, shape $sh_i$ and stroke order $o_i$. We use the coordinate value $(x_i, y_i)$ of $s_{i, 1}$ to represent the location of each stroke $l_i$. In order to disentangle shape information $sh_i$ from its actual location, we use relative instead of absolute coordinates and move the starting point of all strokes to the canvas centre. As per convention, each $sh_i$ point also contains a two-dimensional binary pen state \cite{ha2017neural} -- (1, 0): stroke is being drawn, (0, 1): the end of the stroke, (0, 0): padding points to account for the fact that all strokes have a different number of points.

\keypoint{Sketch-specific encoder.}Our proposed sketch encoder $f_w$, which we name SketchXAINet (``X'' for E$\underline{X}$plainability), first learns to encode $l_i$, $sh_i$ and $o_i$ with different learnable components before fusing them together into a Transformer for final decision. This tailored model design is then ready to undertake the novel explainability task defined later. A full high-level schematic is shown in \cref{fig:framework}. We use a bidirectional LSTM \cite{hochreiter1997long} to extract shape information of each stroke $sh_i$, and one linear layer for location $l_i$ embedding learning. We pre-define the maximum number of strokes allowed and assign a learnable embedding for each order (time) embedding $o_i$. Finally, we sum them all and add one extra \texttt{[CLS]} token before feeding into a transformer encoder \cite{dosovitskiy2020image}. We adopt \texttt{[CLS]} for classification task, optimised under the conventional multi-class cross-entropy loss.

\keypoint{Sketch explainability - SLI.}We introduce a new task for explaining sketch model, that of \textit{Stroke Location Inversion}, SLI. Initiating from replacing each sketch stroke at a random location, SLI explains a sketch classifier through the following hypothesis: to trust a classifier has indeed mastered one concept, a classifier should be able to relocate a group of random strokes towards readable visual imagery that actually corresponds to the underlying concept. SLI thus corresponds to an iterative optimisation problem, aiming to reconfigure strokes locations for increasing recognition confidence. Denoting a sketch composing of $N$ strokes with class label $y$ in bold $\bf{s}$, this process is formulated as:

\begin{equation}
\arg \min_{l_1,\cdots,l_N} \mathcal{L} \left(f_{w}\left(\text{Relocate}(\bf{s})\right), y\right),
\label{eq:LSI}
\end{equation}

\noindent where $\text{Relocate}(\cdot)$ refers to placing the strokes of a given sketch to random locations on a blank canvas.

\keypoint{In connection to counterfactual \& latent optimisation.}At first glimpse, SLI draws considerable similarity to counterfactual explanation -- finding input variations that lead to complete change of prediction outcomes. We adapt this definition under our context with a slight modification to its original formulation \cite{wachter2017counterfactual}:

\begin{equation}
\arg \min_{l_1,\cdots,l_N} \mathcal{L} \left(f_{w}\left(\bf{s'}\right), {y'}\right)+d\left(\bf{s}, \bf{s'}\right),
\label{eq:counterfactual}
\end{equation}

\noindent where ${y'}$ denotes another label different from $y$, $d(\cdot)$ is some distance measure and can be a simple sum of location difference here. The advantage of SLI becomes evident under such comparison. Unlike the counterfactual approach restricted by a local input search space, SLI enjoys a much bigger flexibility with each time explaining a different facet of fact through random replacements of $\bf{s}$. SLI is also connected to latent optimisation, a technique extensively explored in GAN literature \cite{xia2021gan}. If we dissect $f_w$ into $f_l\;\text{(location-relevant component)} \circ f_{w\setminus l}$\;\text{(location-irrelevant component)} and draw an analogy to the latent vector $z$ and generator $G(\cdot)$ in GAN language respectively, this becomes a standard GAN inversion problem. The difference is instead of traversing along the non-interpretable $z$ space, $f_w$ is interpretable in nature with each update dictating the direction and pace of the next sketch stroke movement.

\keypoint{Formal Definition.}We now define two types of SLI tasks, where stroke relocation is leveraged as a gateway to explaining a sketch classifier. \textit{Recovery:} During the recovery task, we randomise the locations of all strokes and only keep their shapes. We specify the target label $y$ as the original sketch label and use \cref{eq:LSI} to optimise ($l_{1},\cdots,l_{N}$).
We visualise the entire optimisation process to understand the inner workings of the classifier. \textit{Transfer:} For the transfer task, we keep stroke shapes and locations intact, while specifying the target label $y$ as a different category to that of the input sketch. We use this setup to build cross-category understandings.

\section{Experiments} \label{sec:experiment}
\subsection{Experimental Settings}
We adopt the QuickDraw dataset \cite{ha2017neural} to train $f_w$, which contains 345 object categories with 75K sketches each. Following convention the 75K sketches are divided into training, validation and testing sets with size of 70K, 2.5K and 2.5K, respectively. For the analysis of generated explanations by \LSI, we randomly select 30 categories. We compare our model with a variety of sketch recognition models: CNN-based \cite{resnet, yu2015sketch}, hybrid-based \cite{sketchmate, yang2021sketchaa, li2018sketch} and Transformer variants \cite{dosovitskiy2020image, liu2021swin, sketchformer}. We use the same learning rate of $0.00001$, \texttt{Adam} optimiser \cite{adam2014kingma}, and 20 epochs for all methods.
All experiments of this stage are run on 5 NVIDIA 3090 GPUs with a batch size of 100 per GPU. For better SLI training  stability, we use \texttt{gradient clip} \cite{pascanu2013difficulty}, \texttt{CosineAnnealingLR} scheduler \cite{loshchilov2016sgdr} and \texttt{SGD} optimiser without momentum to limit the distance a stroke can move.

\subsection{Main Results}

\keypoint{SLI achieves SoTA sketch recognition.}We use top-1 classification accuracy to assess the sketch recognition task. \cref{tab:evaluation_performance} shows performance comparison between all selected models and ours. We include all five major sketch recognition works in contemporary time, Sketch-a-Net \cite{yu2015sketch}, SketchMate \cite{sketchmate}, SketchAA \cite{yang2021sketchaa}, SketchFormer \cite{sketchformer} and Sketch-R2CNN \cite{li2018sketch} and find out Sketch-R2NN has significant edges over others. We also experiment with not sketch-specific but more mainstream vision representation learning architecture, Vision Transformer (ViT) \cite{dosovitskiy2020image} and its more advanced variant Swin Transformer \cite{liu2021swin}. Both however is only on par to SketchFormer, a Transformer-based framework on point, other than patch pixel embedding. SketchXAINet demonstrates that Transformer \textit{can} outperform CNN (Sketch-R2CNN with ResNet-101) on sketch recognition tasks. We achieve a new state-of-the-art sketch recognition performance, improving on all prior arts. We also conduct controlled study to verify the relative importance of ech component in our decomposed stroke representation. Without surprise, the shape feature plays a major role while the order information is the least important.

\begin{table}[t]
\centering
    \begin{adjustbox}{max width=\linewidth}
        \begin{tabular}{l c c}
        \toprule
         {\textbf{Methods}} & {\textbf{Acc.} (\%)} & {\textbf{Params}} \\ \midrule
        ResNet-50 \cite{resnet} & 78.76 & 24.2 \\
        Sketch-a-Net \cite{yu2015sketch} & 68.71 & \underline{8.5} \\
        SketchMate \cite{sketchmate} & 80.51 & 64.7 \\
        ViT-Base \cite{dosovitskiy2020image} & 77.90 & 86.6 \\
        Swin-Base \cite{liu2021swin} & 78.71 & 87.8 \\
        SketchFormer \cite{sketchformer} & 78.34 & 13.1 \\
        SketchAA \cite{yang2021sketchaa} & 81.51 & 26.7 \\
        Sketch-R2CNN \cite{li2018sketch} (ResNet-50) & 84.81 & 32.7 \\
        Sketch-R2CNN \cite{li2018sketch} (ResNet-101) & 85.30 & 51.7 \\
        \noalign{\smallskip}\cdashline{1-3}\noalign{\smallskip}
        SketchXAINet-Tiny (No Shape) & 31.04 & - \\
        SketchXAINet-Tiny (No Location) & 81.41 & - \\
        SketchXAINet-Tiny (No Order) & 83.66 & - \\
        \noalign{\smallskip}\cdashline{1-3}\noalign{\smallskip}
        SketchXAINet-Tiny & \underline{86.10} & \textbf{6.1} \\
        \textbf{SketchXAINet-Base} & \textbf{87.21} & 91.7 \\
        \bottomrule
        \end{tabular}
    \end{adjustbox}
    \caption{Recognition accuracy (\%) and parameters (million) of different methods on 345 categories of QuickDraw \cite{ha2017neural} dataset. Sketch-R2CNN is the previous SoTA. \textbf{Bold} and \underline{underline} denote the best and the second best method. -Base / -Tiny follow the architecture setting in the original ViT work.}
    \label{tab:evaluation_performance}
\end{table}

\begin{figure*} [hbtp]
    \centering
    \includegraphics[width=.95\linewidth]{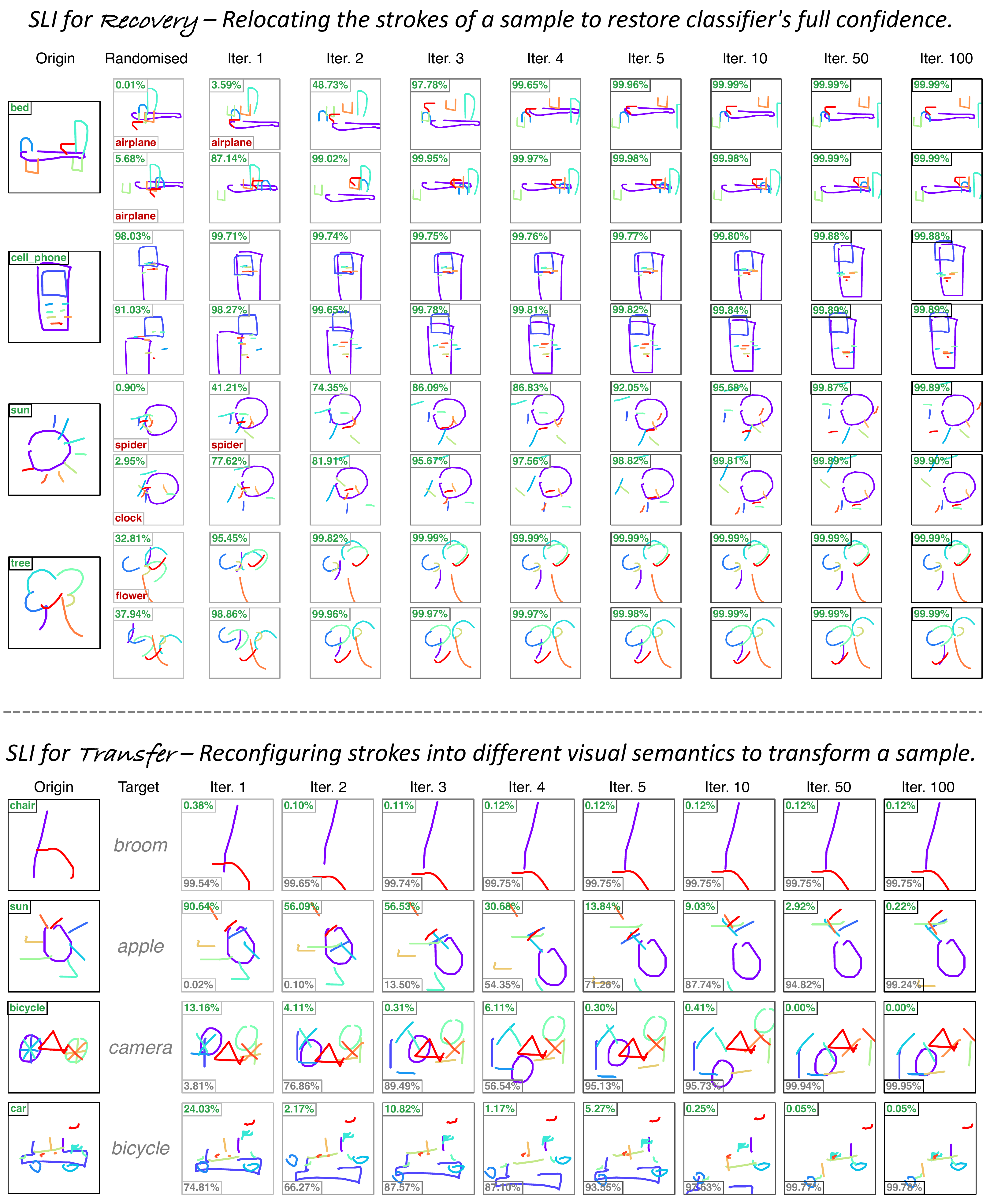}
    \caption{
    \textbf{SLI explains SketchXAINet in Recovery and Transfer tasks.} Here we show the visualisations of the 100 optimisation steps of SLI (Eq.~\ref{eq:LSI}). Origin refers to a free-hand sketch sampled from the QuickDraw dataset, where in recovery we randomise its constituent strokes to form different explainable inputs, and in transfer, we keep it intact but leverage it to explain a classifier of the different target category.
    The number in the top-left corner (the bottom-left corner when present) indicates model confidence in the current sketch to belong to the original label (to the new counterfactual label).
    We use bounding boxes with gradient colours (from light grey to black) to highlight the progressive nature of SLI.
    }
    \label{fig:recovery}
\end{figure*}

\keypoint{SLI provides probe for understanding deep classifier.}
Fig. \ref{fig:recovery} shows the generated visual explanations with SLI taking effect in both recovery and transfer tasks. We first analyse the recovery results with the following observations: i) despite the recovered sketches are often visually different from the original inputs, they reveal the essential category-specific semantics for viewers to interpret, and in turn, build their \textit{own explainability} on how trustworthy the current classifier prediction is. For instance, in the \texttt{[sun]} case, the classifier learns the concept of light by trying to relocate random clustered strokes back to the surroundings of the circle. It is also a bit surprising to see in the first row of \texttt{[tree]} case that the classifier has fostered some fine-grained understanding -- mainly relocating one single but crucial stroke can achieve the transition from the flower stem to the tree trunk. ii) The iteration steps to which optimisation converges vary across samples and randomised starting points, with 100 iterations being a generous enough budget for all scenarios and only taking a few seconds on a modern GPU. iii) SLI helps to identify model flaws. In the \texttt{[cell\_phone]} example, the classifier seems to have not learned a solid correct spatial composition of its components. For \texttt{[tree]}, while the recognition confidence climbs to $95.45\%$ from $32.81\%$ after the $1^{st}$ iteration, we fail to observe convincing visual effect change accordingly, indicating some less explainable model behaviour. iv) Back to the transfer task, we can see that the generated explanation becomes less effective but is still partially understandable. SLI is able to put strokes at the right place representing the \textit{just right abstraction} of visual semantics. The seat stroke of a \texttt{[chair]} turns into the head of a \texttt{[broom]} and the \texttt{[bicycle]} is totally anatomised to resemble the looking of a \texttt{[camera]}. Downsides of a classifier are also implied where inverting a \texttt{[sun]} into \texttt{[apple]} reveals the vulnerability of the apple classifier under a pineapple attack. In summary, SLI provides an interpretable tool to visually probe into the functioning of a sketch classifier and enable various AI explainability projections.

\begin{figure*}[t]
    \centering
    \includegraphics[width=\linewidth]{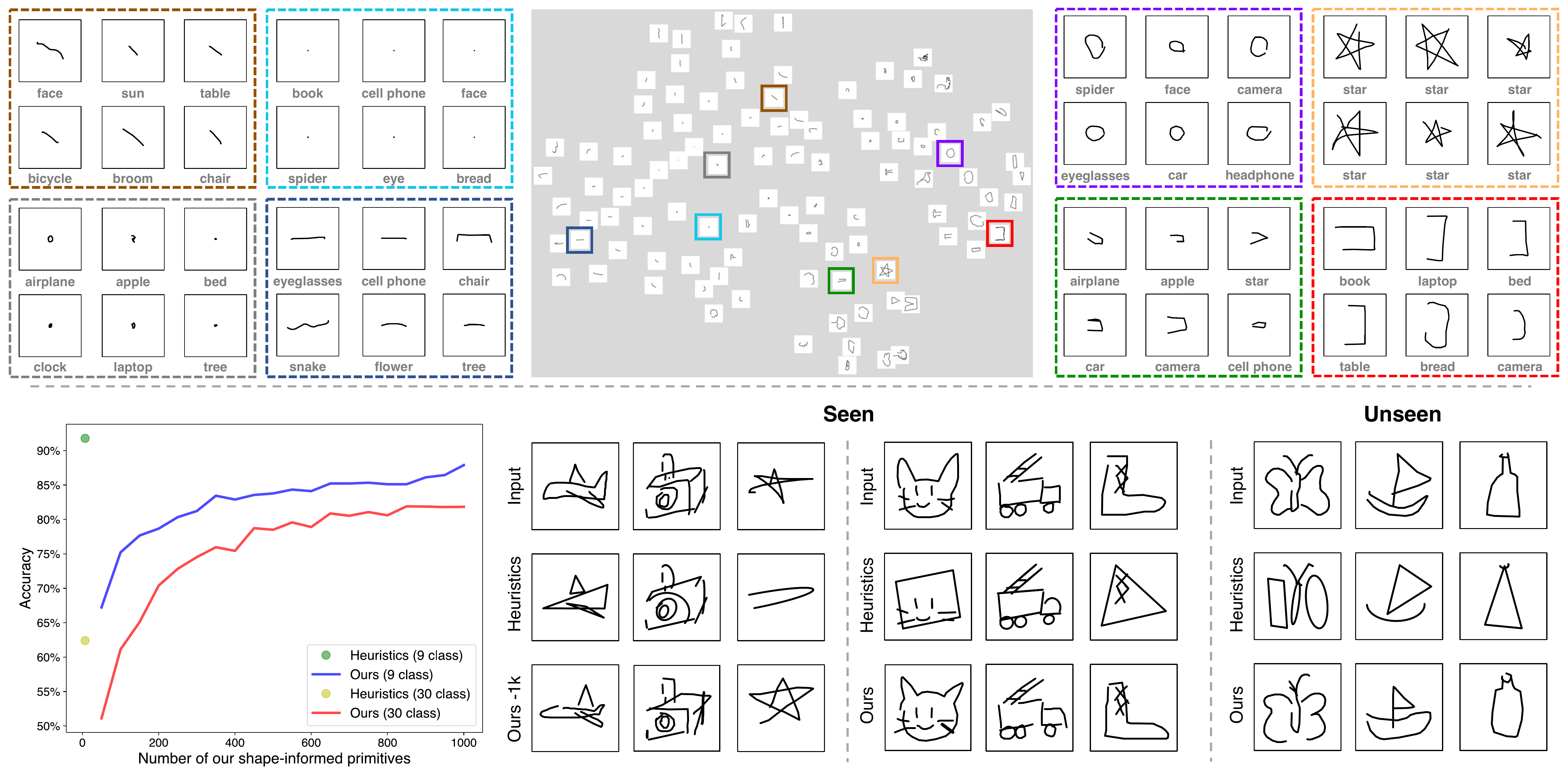}
    \caption{\textbf{Analysis on shape embedding.} Top: t-SNE visualisation on 100 stroke primitives across 30 sketch categories. Strokes with similar semantics are grouped together regardless of the original categories sourced from. Bottom: we compare our learned stroke primitives with \cite{alaniz2022primitives}, where 7 stroke primitives are heuristically pre-defined and their efficacy to reconstruct a sketch (\ie,  replace any stroke with a primitive) is evaluated on a carefully curated 9-class setting. The table shows the method largely fails when extending the evaluation to a more open-world setting of 30 classes. Ours can not only deal with less regularised sketches from seen classes (\eg, star), but also generalises well to unseen cases.}
    \label{fig:shape_embeddings}
\end{figure*}

\subsection{On Stroke Shape Embedding}

\begin{figure}[th]
    \centering
    \includegraphics[width=\linewidth]{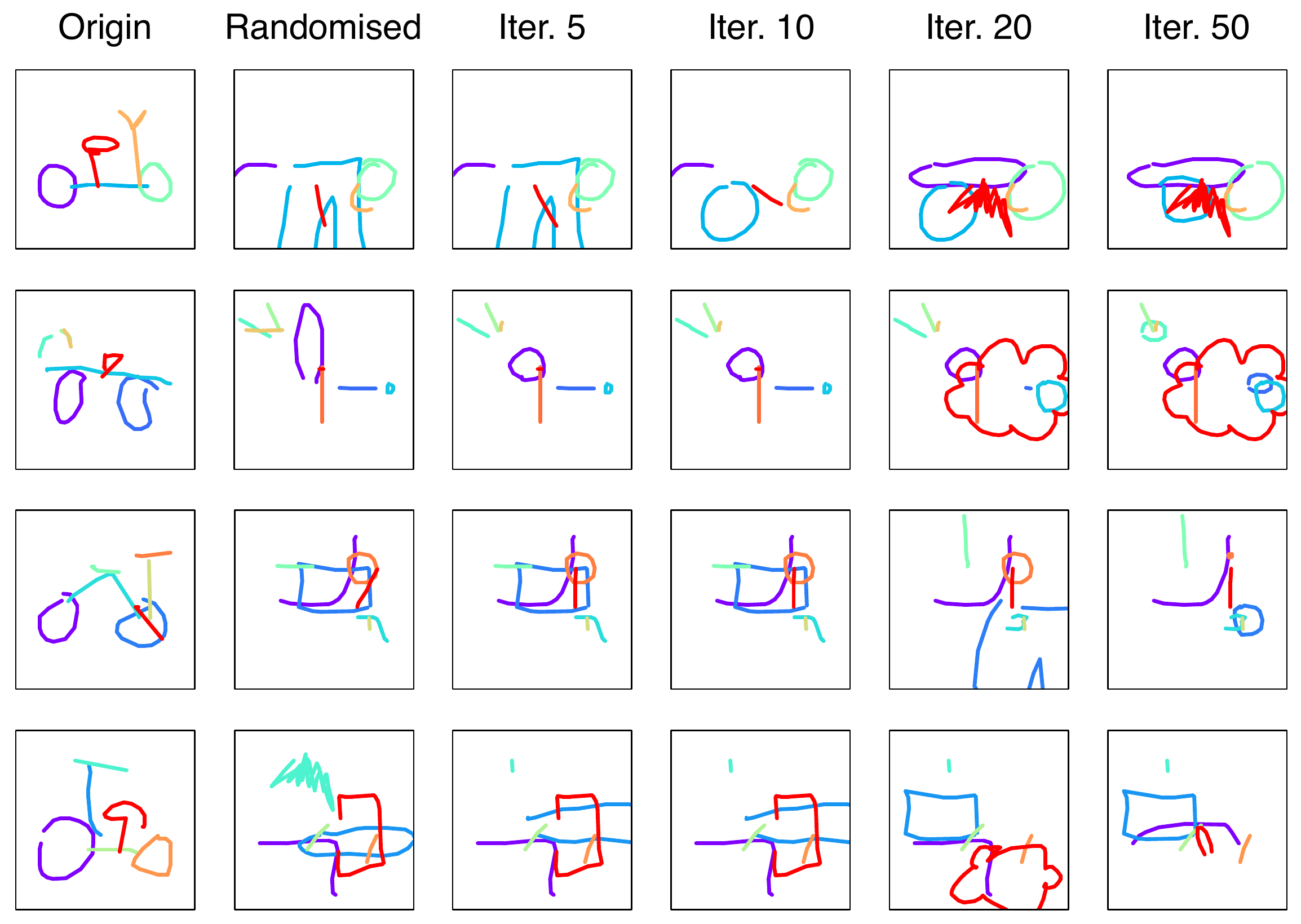}
    \caption{\textbf{Shape, not location, Inversion.} With automatically generated stroke primitives, we can now proceed inversion tasks on stroke shapes, just like how we do for locations -- updates on high-dimensional shape embedding can be now visualised to changes of shape primitives if that update becomes significant enough. We however fail to identify explainable factors in such inversion.}
    \label{fig:si}
\end{figure}

To analyse our learned shape embedding\footnote{Analysis on order embedding can be found in the supplementary.}, we conduct t-SNE \cite{van2008visualizing} across the strokes of the selected samples from all sketch categories and run K-means on their reduced dimensions. We simply define each cluster centroid as the stroke sample (during training) closest to and see that as the \textit{representative stroke primitive} of all stroke samples belonging to the same centroid. A natural outcome is that the larger centroid numbers we set in K-means, the finer primitives incorporating more diverse drawing styles are expected. The first row of Fig.~\ref{fig:shape_embeddings} shows the t-SNE clustering results with 100 centroids on 30 sketch categories and confirms the shape embedding has formed semantics understanding to group visually similar strokes together regardless of the original category they come from -- see how dots with different hollow types are well recognised by the embedding. For more quantitative evaluation, we replace all strokes of a sketch sample with their primitives and feed them into SketchXAINet for classification. Comparing with the results reported in the past work \cite{alaniz2022primitives} which manually define a fixed set of heuristics-based shape primitives (line, arc, square, circle, triangle, U-shape, L-shape), our learning-based method is flexible in how a stroke is to be abstracted and how to trade-off recognition at the whole sketch level therein. We demonstrate the comparison in the bottom row of Fig.~\ref{fig:shape_embeddings}. Apart from the 9-class setting from \cite{alaniz2022primitives} that specifically choose certain classes with visual semantics biased to their analysis (\eg, round-shaped silhouette), \cite{alaniz2022primitives} mostly fails under more open setting, with recognition accuracy plummeting from 91.8\% to 62.4\% in 30-class setting and complete reconstruction failure for less regularised sketch samples (\eg, shoe, star). Finally, with learned stroke primitives, we can now try to conduct shape, rather than stroke inversion explainability task by modifying Eq.~\ref{eq:LSI} to optimise $sh_1, sh_2,..., sh_n$ instead. After each gradient descent, we replace the updated shape embeddings with their closest primitives and use them as initialisation for the next step. Examples in Fig~.\ref{fig:si} show that shape inversion hardly delivers any explainable outcome and implicitly justifies our location inversion choice.

\begin{figure}[t]
    \centering
    \includegraphics[width=\linewidth]{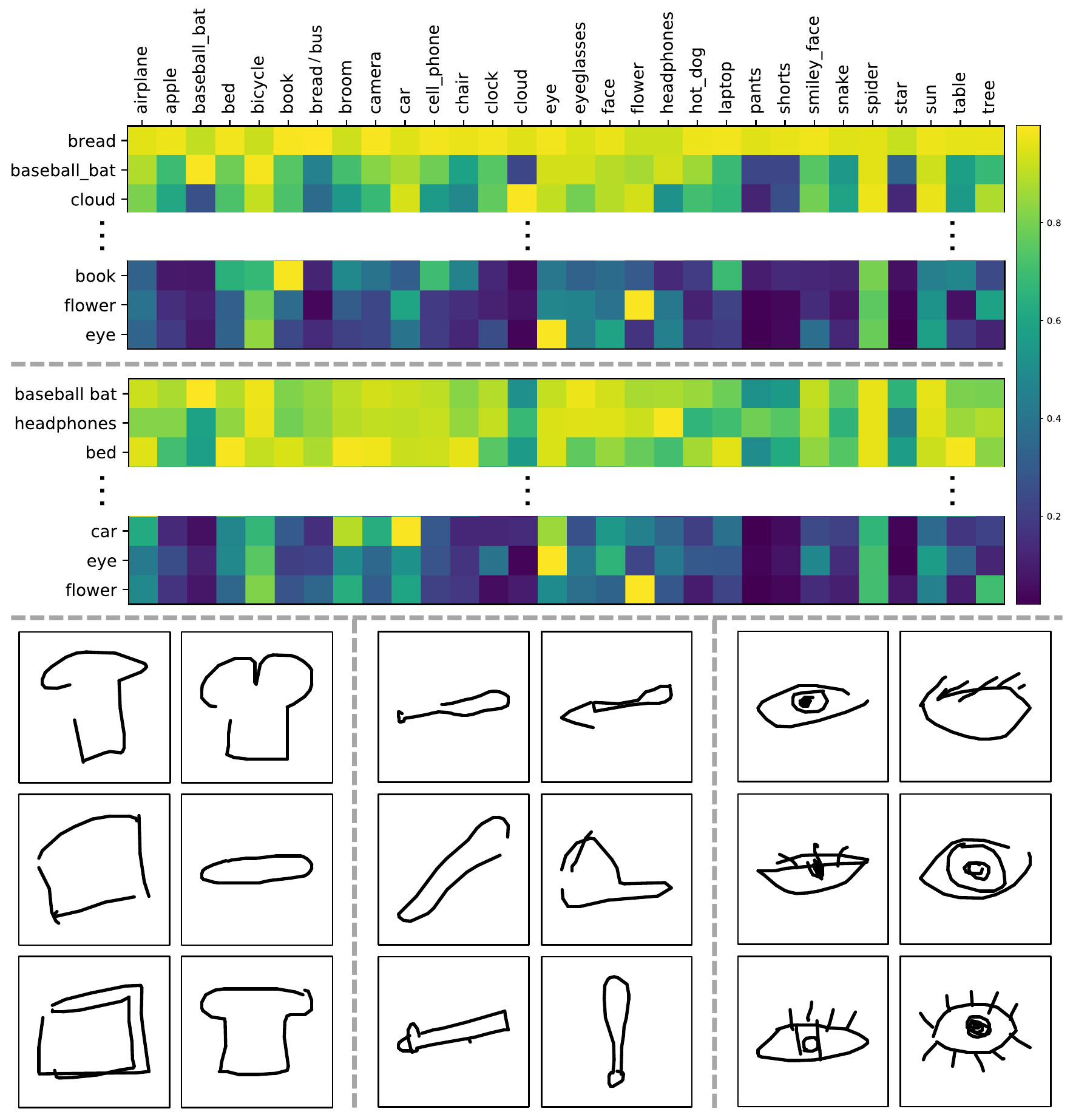}
    \caption{\textbf{SLI exposes dataset bias.} Top: we apply SLI on transfer tasks between every two categories out of a total of 30 and observe all sketch samples regardless of the origin can be successfully transferred to \texttt{[bread]} (left). To confirm, we exclude \texttt{[bread]} and replace it with a new category \texttt{[bus]} and this time all sketches transfer to \texttt{[baseball\_bat]}. Bottom: we showcase some samples of three QuickDraw categories, \texttt{[bread]}, \texttt{[baseball\_bat]}, \texttt{[eye]}, which yields an explanation to the said phenomenon. More details in text.}
    \label{fig:discussion}
    \vspace{0.2cm}
\end{figure}

\begin{figure}[t]
    \centering
    \includegraphics[width=\linewidth]{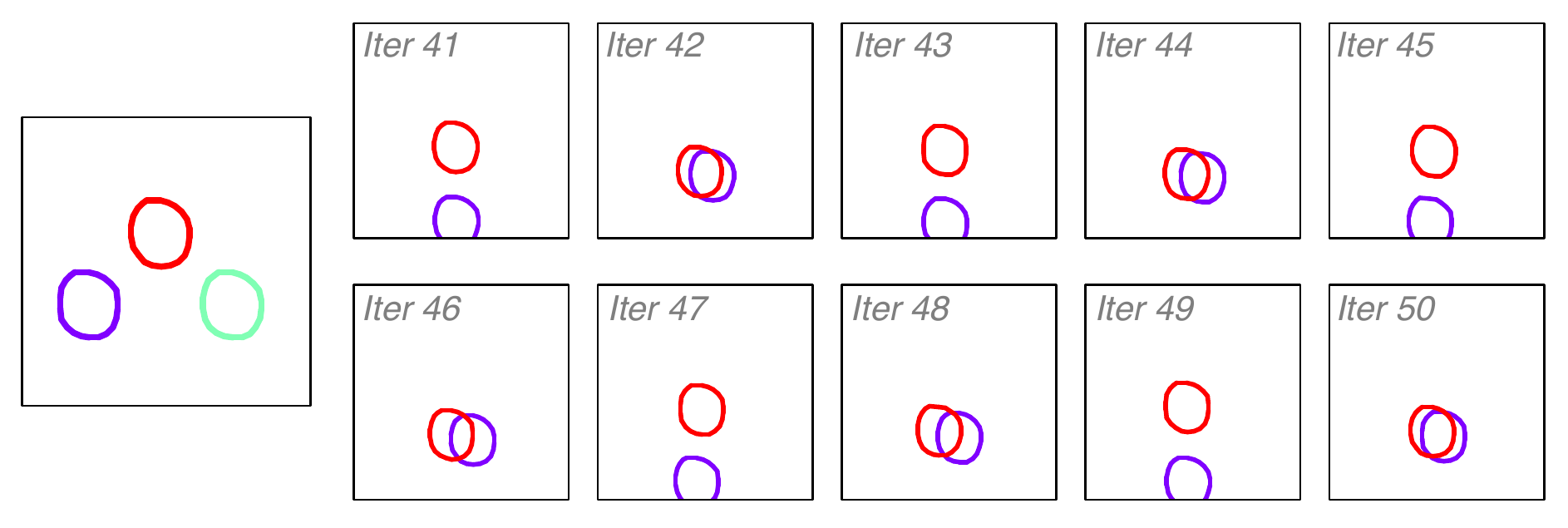}
    \caption{\textbf{Limitation.} SLI relies on gradient descent and thus inherits its weakness. Here we demonstrate with a simple \texttt{sun} transfer task how optimisation is trapped in local optima.}
    \label{fig:limit}
\end{figure}

\section{Discussion}

\keypoint{Explaining dataset bias with \SLI{}}
In our transfer explainability setting, we showed that by relocating the strokes and in some cases removing the strokes from the canvas (moving them out of the canvas bounding box) we can transfer a sketch from category A to category B.
Here, we conduct an additional experiment.
We sample $1000$ sketches for each of the $30$ training categories and apply a transfer task for each pair of sketches.
In the top part of Fig.~\ref{fig:discussion}, we visualise as a heat map \footnote{The full heat map can be found in the supplementary.} the average recognition confidence values to belong to the target category of sketches transferred from one category to another (The first row indicates that when the target category is \texttt{[bread]}, the average confidence of the samples of 30 categories is the highest, and so forth).
We find that for almost all sketch categories the average confidence is high for a transfer to a sketch of \texttt{[bread]}.
Then, we naturally ask the question of how this behaviour can be explained.
We start by looking at the example of the sketches from the \texttt{[bread]} category.
In Fig.~\ref{fig:discussion} bottom, we show sketch samples from the QuickDraw dataset for bread sketches\footnote{\url{https://quickdraw.withgoogle.com/data/bread}}, we can see that many look like something else, e.g.~a \texttt{[shirt]}.
Our \SLI{} task allowed us to find a category for which sketches are ambiguous with respect to an assigned category.
The next category with high average confidence of the transfer task, \texttt{[baseball\_bat]}, also contains many ambiguous sketches, for example, resembling a \texttt{[knife]}.
We also show the \texttt{[eye]} sketches, which we find to be the category hardest to transfer to. We can see that all sketches do look like eyes.
Therefore, we can see how our \SLI{} task can help to identify categories for which humans struggle to produce easily recognisable sketches. Such dataset bias needs to be taken into account when training deep models.
To conclude, this pilot study provides further insights into how \SLI{} contributes towards explainability.

\keypoint{Limitation.} SLI is based on gradient descent and therefore inherits its limitations: SLI can be susceptible to local optima by oscillating around stroke location and not progressing further. We exemplify this in Fig.~\ref{fig:limit} where we use three circles to explain the sun concept. The expectation is then that two circles will be driven away off the canvas and one circle left.
In practice, however, one circle is driven away and two circles are trapped in a tug-of-war. Solutions to alleviate this issue can be inspired by the optimisation literature, \eg, look ahead optimiser \cite{zhang2019lookahead} is designed to break the optimisation deadlock by maintaining two sets of fast and slow weights.

\section{Conclusion}

Sketches form a great data modality for explainability research because of their inherent ``human-centred'' nature. We started our journey by first identifying strokes as the basis for explanation. We then introduced SketchXAINet to encode the three innate properties of sketch strokes: shape, location, and order. Leveraging this encoder, we propose the first sketch-specific explainability task, that of stroke location inversion (SLI).
Compared to your typical static explanations (\eg, saliency map), SLI is a dynamic process that explains the credibility of a sketch model by examining its ability to relocate randomly reshuffled strokes to reconstruct a sketch given a category. We attest to the efficacy of SLI with extensive analysis and contribute a new SoTA sketch recognition model as a by-product.
Last but not least, we repeat that this is only the very first stab, yet at what we believe to be a very important and interesting area for XAI.

{\small
\bibliographystyle{ieee_fullname}
\bibliography{arxiv.bib}
}

\clearpage
\appendix

\onecolumn
\section{Existing explainable methods applied to sketches}

\begin{figure} [h]
    \centering
    \includegraphics[width=0.96\linewidth]{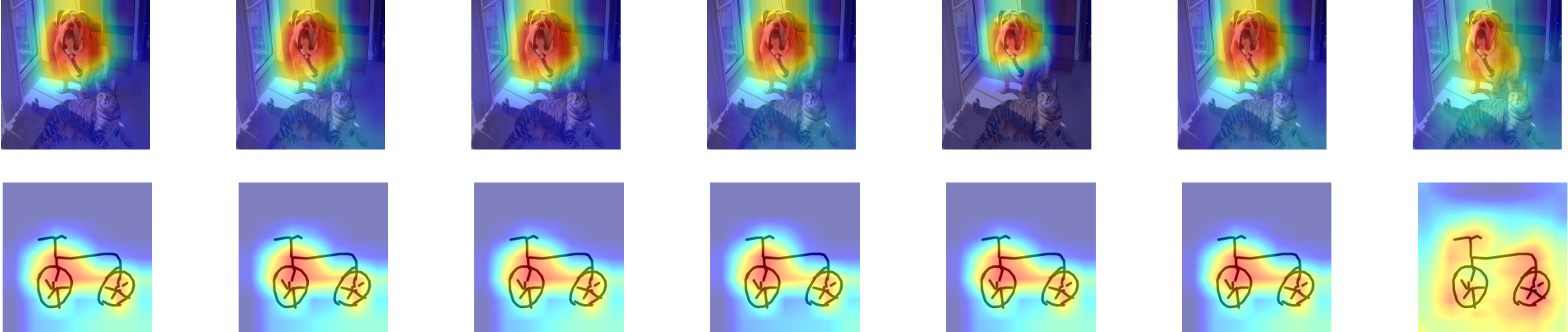}
    \vspace{1em}
    \caption{
    \textbf{Comparison of image-based explainable methods applied to images and sketches, which is indicated above each column.}
    }
    \label{fig:explainable_cv}
\end{figure}

\cref{fig:explainable_cv} visualises several explainable methods applied to images and sketches. It can be observed that image-based methods are more helpful on images than on sketches: Sketches are sparse and to some extent already represent salient features of a depicted object, therefore the image-based explainable methods tend to highlight most of the sketch.

\vspace{1em}

\begin{table*}[!ht]
\adjustbox{width=\linewidth}{
    \begin{tabular}{P{2.5cm} P{3cm} P{2.5cm} P{2.8cm} P{3.1cm} P{2.5cm} P{2.5cm}}
    \toprule
    \textbf{GradCAM} \cite{selvaraju2016grad} & \textbf{GradCAM++} \cite{chattopadhay2018grad} & \textbf{XGradCAM} \cite{fu2020axiombased} & \textbf{AblationCAM} \cite{ramaswamy2020ablation} & \textbf{EigenGradCAM} \cite{muhammad2020eigen} & \textbf{LayerCAM} \cite{jiang2021layercam} & \textbf{FullCAM} \cite{srinivas2019full} \\
    \midrule
    \end{tabular}
}
\end{table*}

\vspace{-1.75em}

\vspace{1em}

\begin{figure}[h!]
    \centering
    \includegraphics[width=0.75\linewidth]{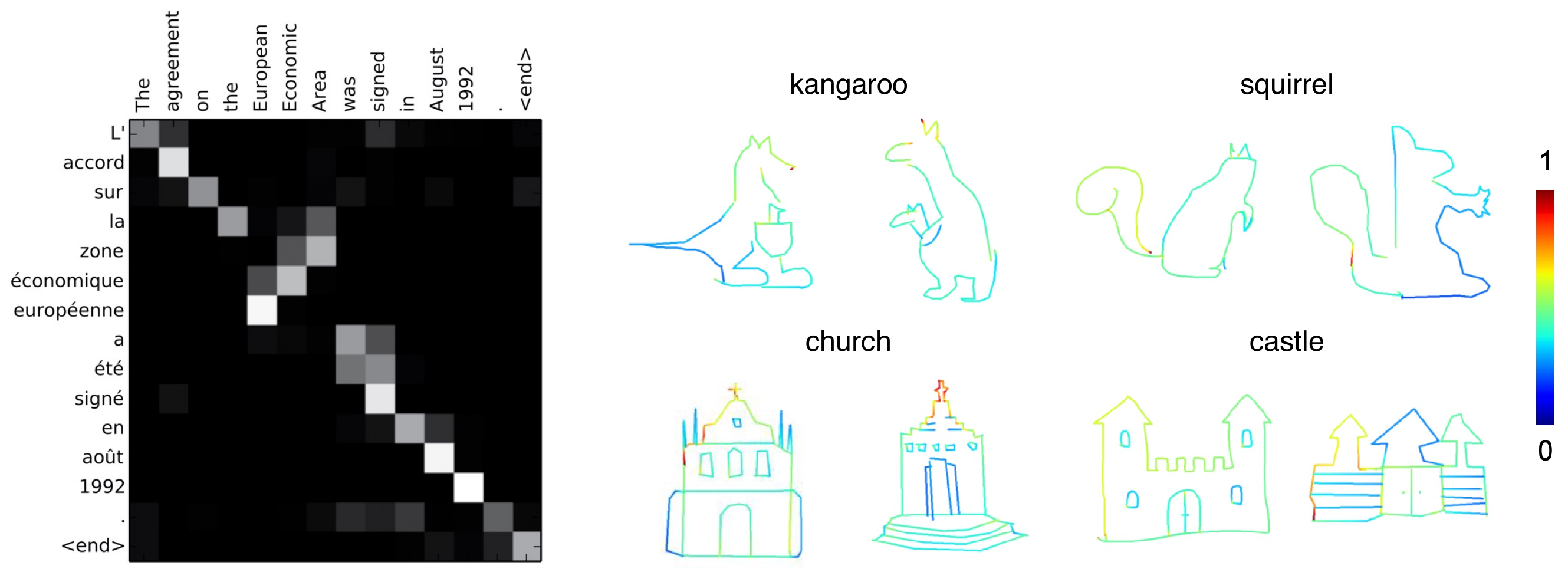}
    \vspace{1em}
    \caption{
    \textbf{Comparison of some text-based explainable methods applied to text and sketches.} (\textbf{Left}) The classical visualisation of alignment between words in Neural Machine Translation \cite{bahdanau2014neural}. (\textbf{Right}) Colour-coded attention maps produced by Sketch-R2CNN \cite{li2018sketch}.
    }
    \label{fig:explainable_nlp}
\end{figure}

\cref{fig:explainable_nlp} adopts text-based explainable methods to sketch inputs. \cref{fig:explainable_cv} (left) for the text is informative. In \cref{fig:explainable_cv} (right), we plot attention maps, where the colour of each point indicates its impact on the final classification decision. However, it might be challenging to make any conclusions from these visualisations, as they look rather noisy. For instance, observe how attention for points on the left and right sides in symmetric objects (church, castle) differ. We think that the main reason is that in this case the explainable component is based on points rather than strokes. In contrast, in our work we operate on strokes and their descriptors: shape, location and order. This allows us to design better visualisations for the explainability of sketch classification tasks.

\vspace{1em}

\vspace{1em}

\section{Sketch/strokes representation}
There are two popular formats that are used in deep-learning sketch literate \cite{ha2017neural,muhammad2018learning,lopes2019learned,sketchbert,sketchformer} to represent vector sketches:  ``Stroke-3'' and ``Stroke-5''.
``Stroke-3'' format represents each point as ($\delta x, \delta y, p$), where $\delta x, \delta y$ store point's relative position to the previous point, and $p$ is a binary pen state (0 means drawing and 1 means end of a stroke).
``Stroke-5'' format ($\delta x, \delta y, p_1, p_2, p_3$) has three binary pen states $p_1 - draw$, $p_2 - lift$, $p_3 - end$, which are mutually exclusive.
The important difference with our design is that we separate stroke location from its shape, as described in the main paper. Namely, in our case $\delta x, \delta y$ store relative position to the first point of each stroke.
Then, we use the second pen state to indicate padding points to account for the fact that all strokes have different number of points.
This new representation is key to our SLI settings.
We provide a pseudocode in \cref{alg:comparison}.

\vspace{2em}

\begin{algorithm}
\caption{Pseudocode comparison of the two sketch representation approaches}
\label{alg:comparison}

\definecolor{codeblue}{rgb}{0.25,0.5,0.5}
\lstset{
  backgroundcolor=\color{white},
  basicstyle=\fontsize{7.2pt}{7.2pt}\ttfamily\selectfont,
  columns=fullflexible,
  breaklines=true,
  captionpos=b,
  commentstyle=\fontsize{7.2pt}{7.2pt}\color{codeblue},
  keywordstyle=\fontsize{7.2pt}{7.2pt},
}
\begin{lstlisting}[language=python]
    # Previous approach
    for sketch in sketch_list:
        vector = sketch.vector_data

        # Calculate offset between all points in each sketch
        new_vector[1:] = vector[1:] - vector[0:-1]
        new_vector[0] = vector[0]
        input(new_vector)

    # Our approach
    for sketch in sketch_list:
        for order, stroke in enumerate(sketch):
            vector = stroke.vector_data

            # Calculate offset between all points in each stroke
            new_vector[1:] = vector[1:] - vector[:-1]

            # Save the starting point of each stroke (locations -> global information)
            locations_list.append(vector[0])

            # Reset the starting point
            new_vector[0] = [0, 0]

            # After removing the global information, the rest is shape (shapes -> local information)
            shapes_list.append(new_vector)
            order_list.append(order)

        input(order_list, shapes_list, locations_list)
\end{lstlisting}
\end{algorithm}

\vspace{2em}
\section{Experimental setup and hyperparameters}
\vspace{1em}

\begin{minipage}[b]{0.67\linewidth}

For all ViT experiments, we use the pretrained models from \href{https://huggingface.co/google/vit-base-patch16-224}{ViT of HuggingFace}.
Our SLI task for each sketch involves an optimisation process, where the only learnable parameters are strokes' locations.
As on optimisation is performed \emph{per sketch}, the best learning rate is different for each sample.
To adapt for the diversity of sketches, we use the \texttt{CosineAnnealingLR} scheduler with a maximum learning rate of 10 and a minimum value of 0.00001.
In addition, we use \texttt{gradient clip} \cite{pascanu2013difficulty} to force a maximum horizontal and vertical stroke movement of 0.5, \ie, a quarter of the canvas size.

The 30 categories that we randomly selected for our experiments are \texttt{[airplane]}, \texttt{[apple]}, \texttt{[baseball\_bat]}, \texttt{[bed]}, \texttt{[bicycle]}, \texttt{[book]}, \texttt{[bread]}, \texttt{[broom]}, \texttt{[camera]}, \texttt{[car]}, \texttt{[cell\_phone]}, \texttt{[chair]}, \texttt{[clock]}, \texttt{[cloud]}, \texttt{[eye]}, \texttt{[eyeglasses]}, \texttt{[face]}, \texttt{[flower]}, \texttt{[headphones]}, \texttt{[hot\_dog]}, \texttt{[laptop]}, \texttt{[pants]}, \texttt{[shorts]}, \texttt{[smiley\_face]}, \texttt{[snake]}, \texttt{[spider]}, \texttt{[star]}, \texttt{[sun]}, \texttt{[table]} and \texttt{[tree]}.

\end{minipage}
\hfill
\begin{minipage}[b]{0.25\linewidth}
\begin{figure} [H]
    \centering
    \includegraphics[width=\linewidth]{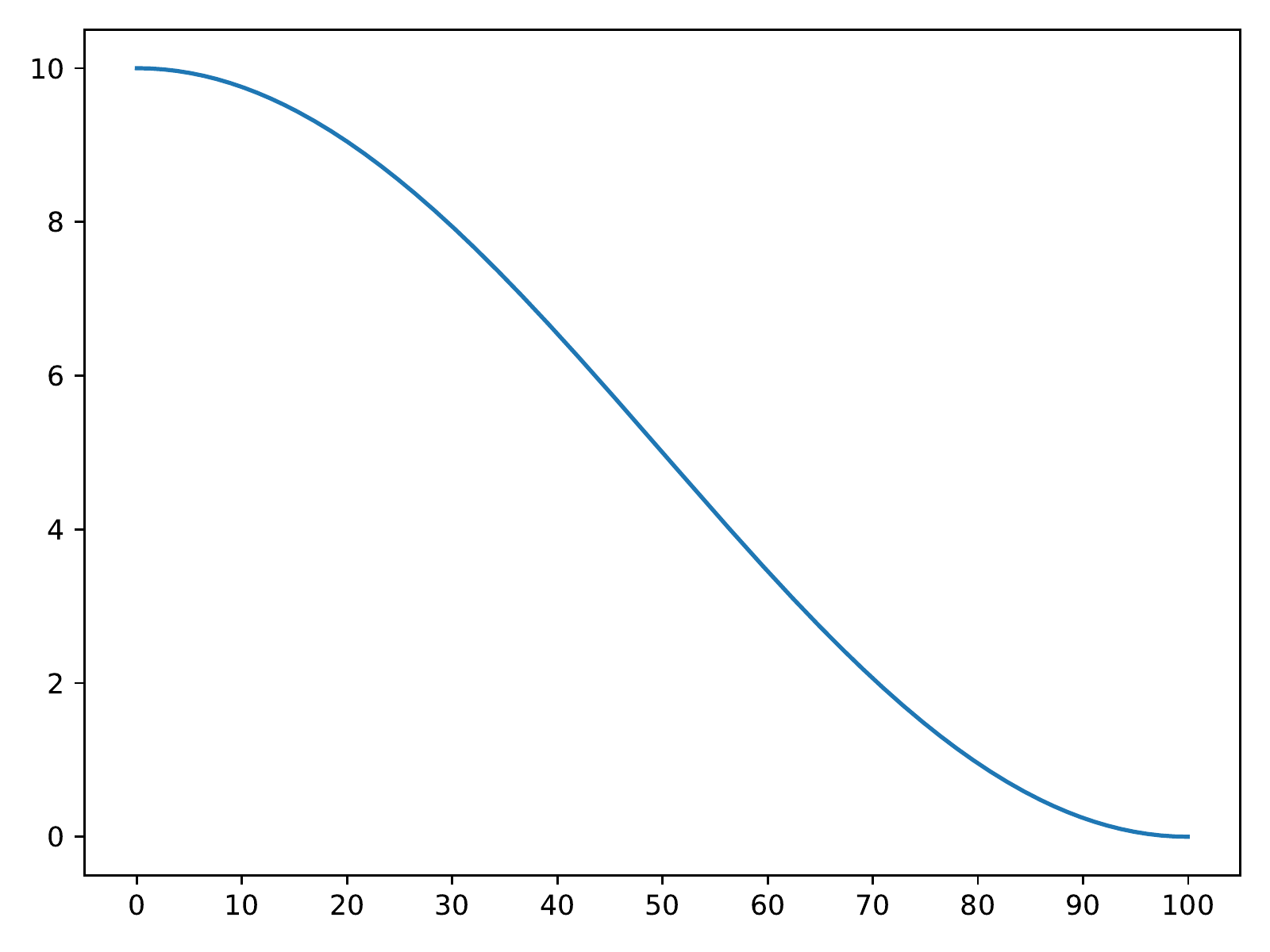}
    \caption{
    The attenuation curve of learning rate over 100 iterations of \texttt{CosineAnnealingLR} scheduler.
    }
    \label{fig:order}
\end{figure}
\end{minipage}

\vspace{2em}
\section{Analysis of stroke order embeddings}
\vspace{1em}

To shed light on the role of stroke order in the decision process of our classifier, we plot stroke order embeddings similarity map, in style of positional embeddings similarity maps of ViT \cite{dosovitskiy2020image}.

\cref{fig:order} right shows the statistic on the number of sketches with a certain number of strokes in the training set.
It can be seen, that over ninety percent of sketches have no more than 10 strokes in the QuickDraw dataset.
Therefore, we plot similarity maps only for the embeddings of the stroke order numbers from 1 to 16.
We break the $16\times 16$ similarity map into $4\times 4$ squares for the visualisation clarity (\cref{fig:order} left).
While the pattern is not too pronounced, it can be seen that the earlier strokes are mostly distinctive from each other, while the further you go the more similar the order embeddings become.
This can be explained by the fact that as different sketches have different number of strokes, the strokes with larger order numbers participate less frequently in model training.
Therefore, the model learns that earlier strokes are more important and their embeddings are more distinguishable from each other.
\vspace{1em}

\begin{figure} [h]
    \centering
    \includegraphics[width=0.9\linewidth]{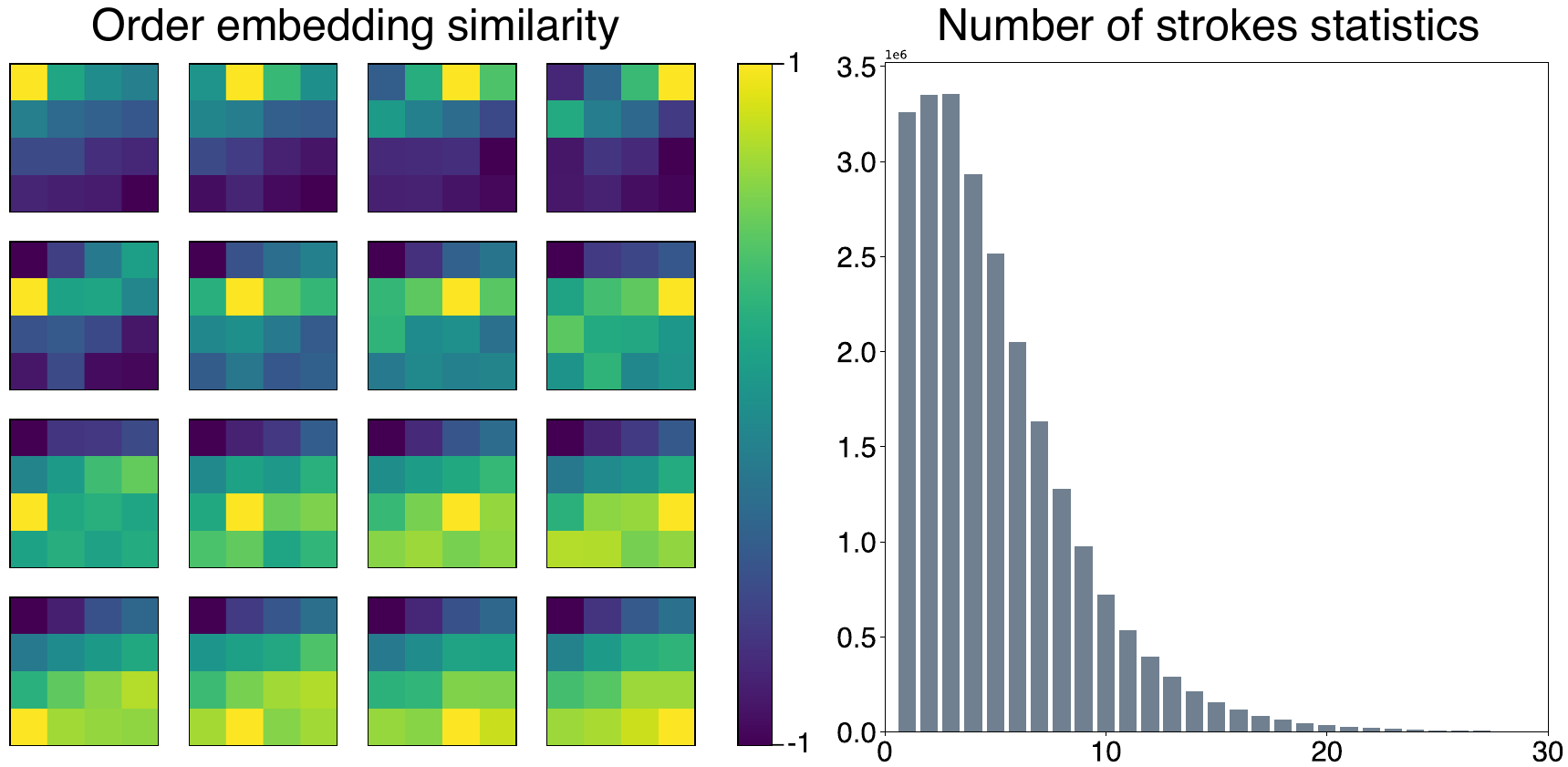}
    \caption{Similarity map of order embeddings and statistics on the number of strokes for all sketches of the training set.}
    \label{fig:order}
    \vspace{2mm}
\end{figure}

\section{Visualisations of shape embeddings}

\begin{figure} [H]
    \centering
    \includegraphics[width=0.85\linewidth]{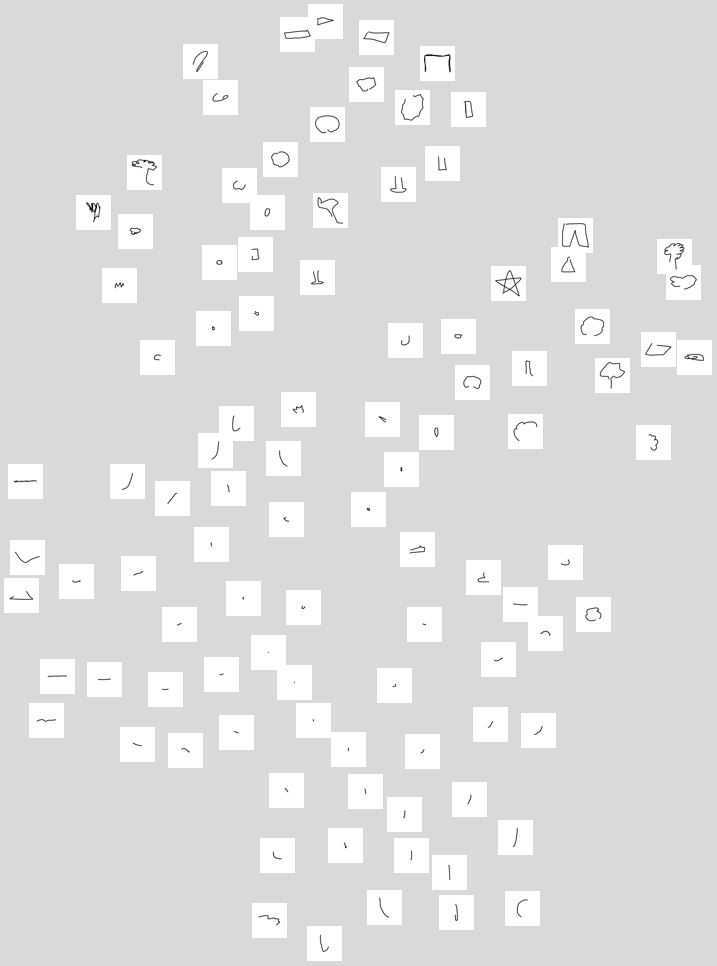}
    \caption{
    \textbf{t-SNE of strokes belonging to 100 clustering centres.} Each of the clustering centres contains hundreds of strokes. In this figure, we select only the nearest stroke to each clustering centre as representative.
    }
    \label{fig:tsne}
\end{figure}

\begin{figure} [H]
    \centering
    \begin{subfigure}[h]{0.4\textwidth}
        \includegraphics[width=\textwidth]{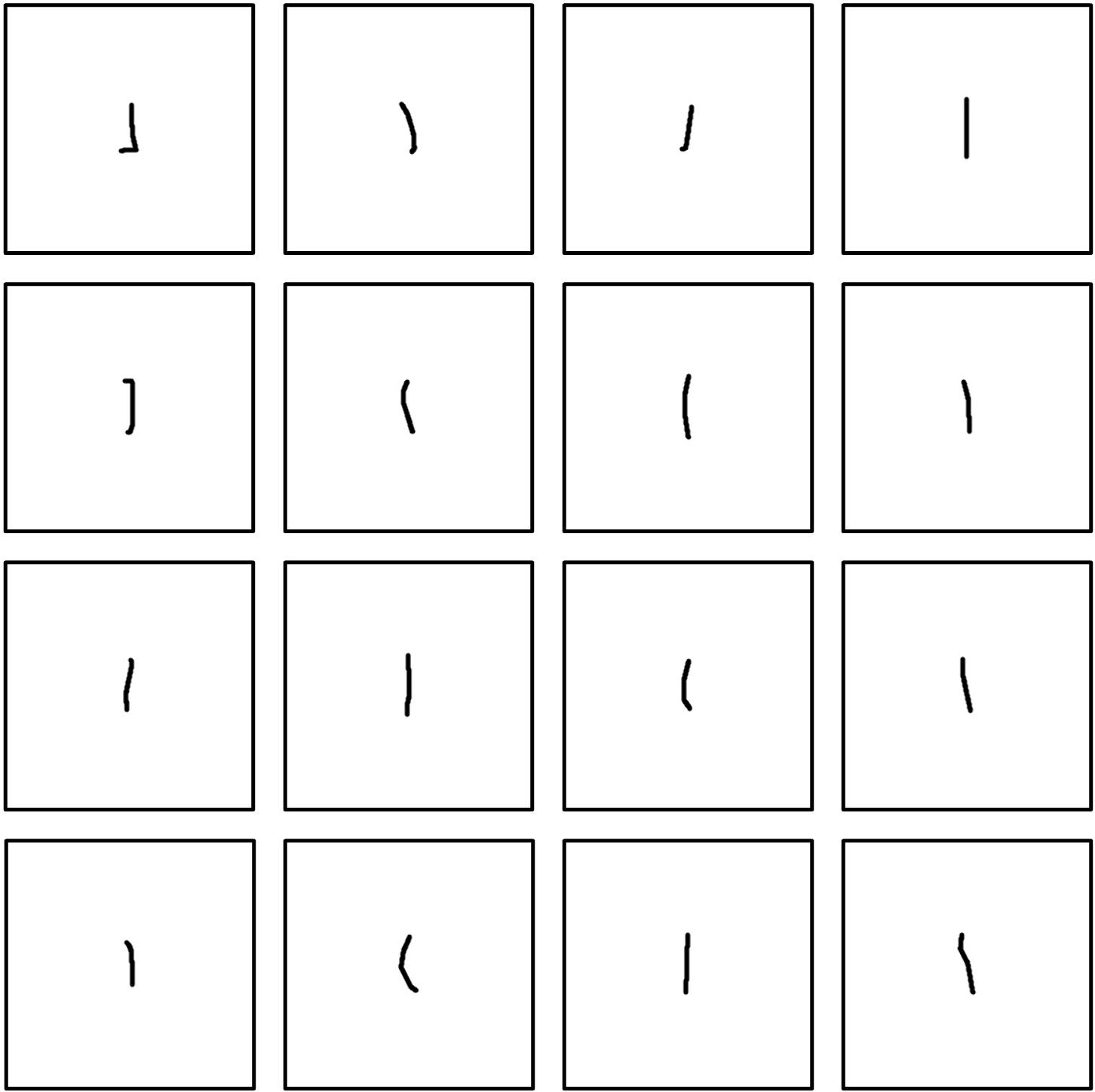}
        \caption{ID=15}
        \vspace{1em}
    \end{subfigure}%
    \quad
    \begin{subfigure}[h]{0.4\textwidth}
        \includegraphics[width=\textwidth]{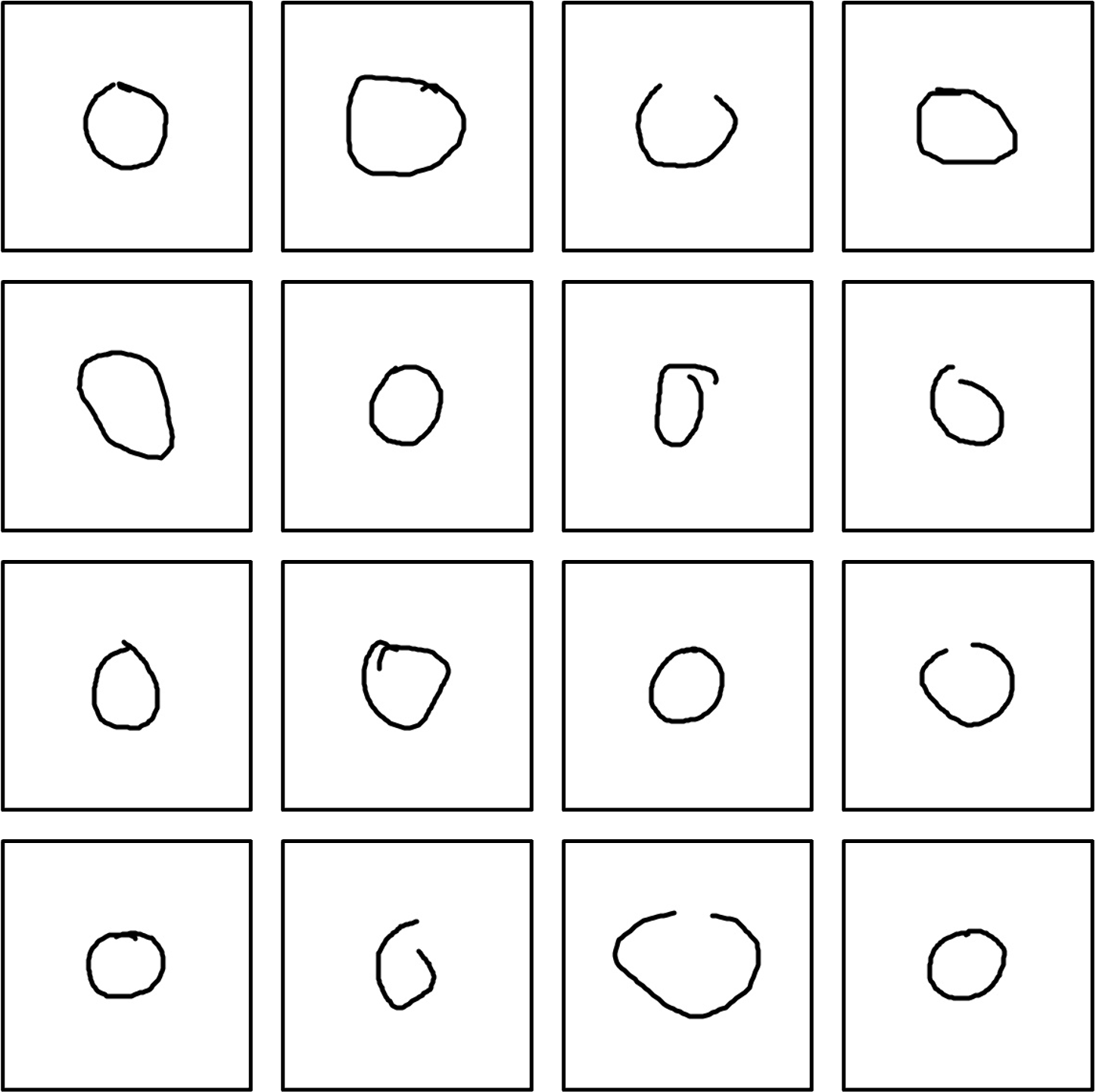}
        \caption{ID=67}
        \vspace{1em}
    \end{subfigure}

    \begin{subfigure}[h]{0.4\textwidth}
        \includegraphics[width=\textwidth]{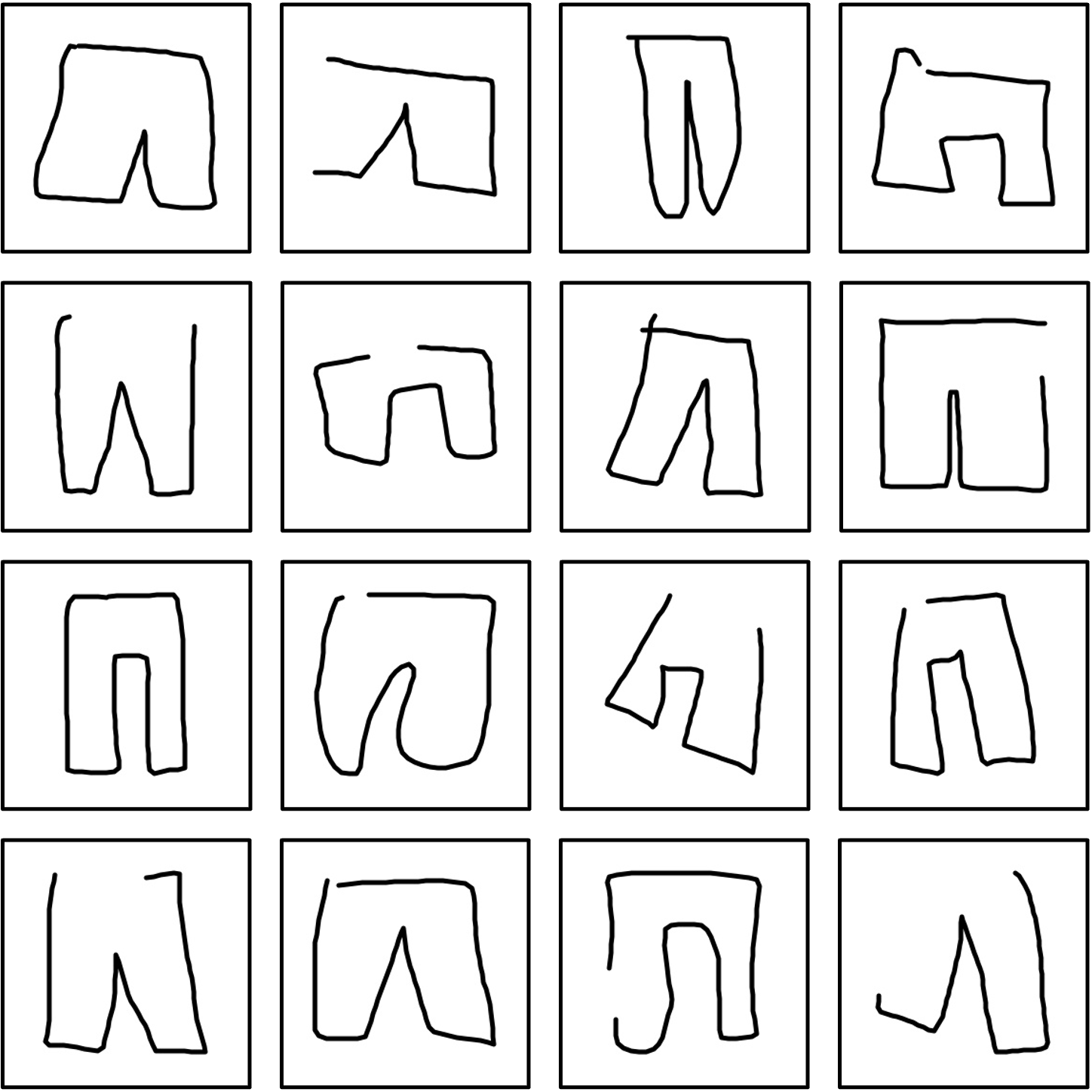}
        \caption{ID=19}
        \vspace{1em}
    \end{subfigure}
    \quad
    \begin{subfigure}[h]{0.4\textwidth}
        \includegraphics[width=\textwidth]{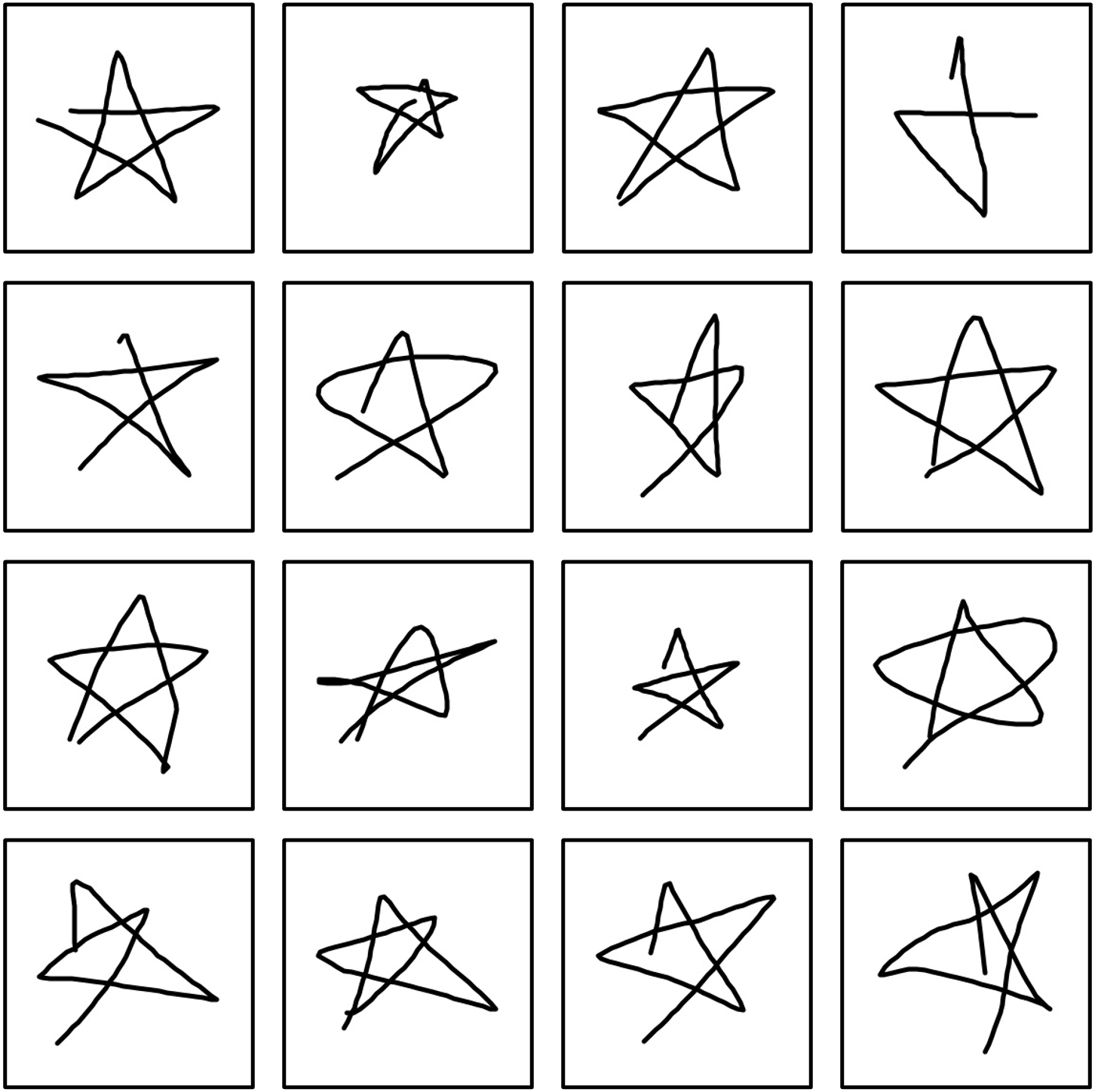}
        \caption{ID=14}
        \vspace{1em}
    \end{subfigure}
    \vspace{1em}
    \caption{\textbf{Visualisations of K-means clustering of shapes.} We randomly select four indices of 100 clustering centres to present in this supplementary material. From the above results we can find that our shape embeddings are learned to group meaningful and consistent semantics of stroke shapes into different indices. Therefore, we can perform stroke replacement tasks based on this property.}\label{fig:codebook}
\end{figure}

\section{Build transfer map based on SLI}

\begin{figure} [H]
    \centering
    \vspace{3em}
    \includegraphics[width=\linewidth]{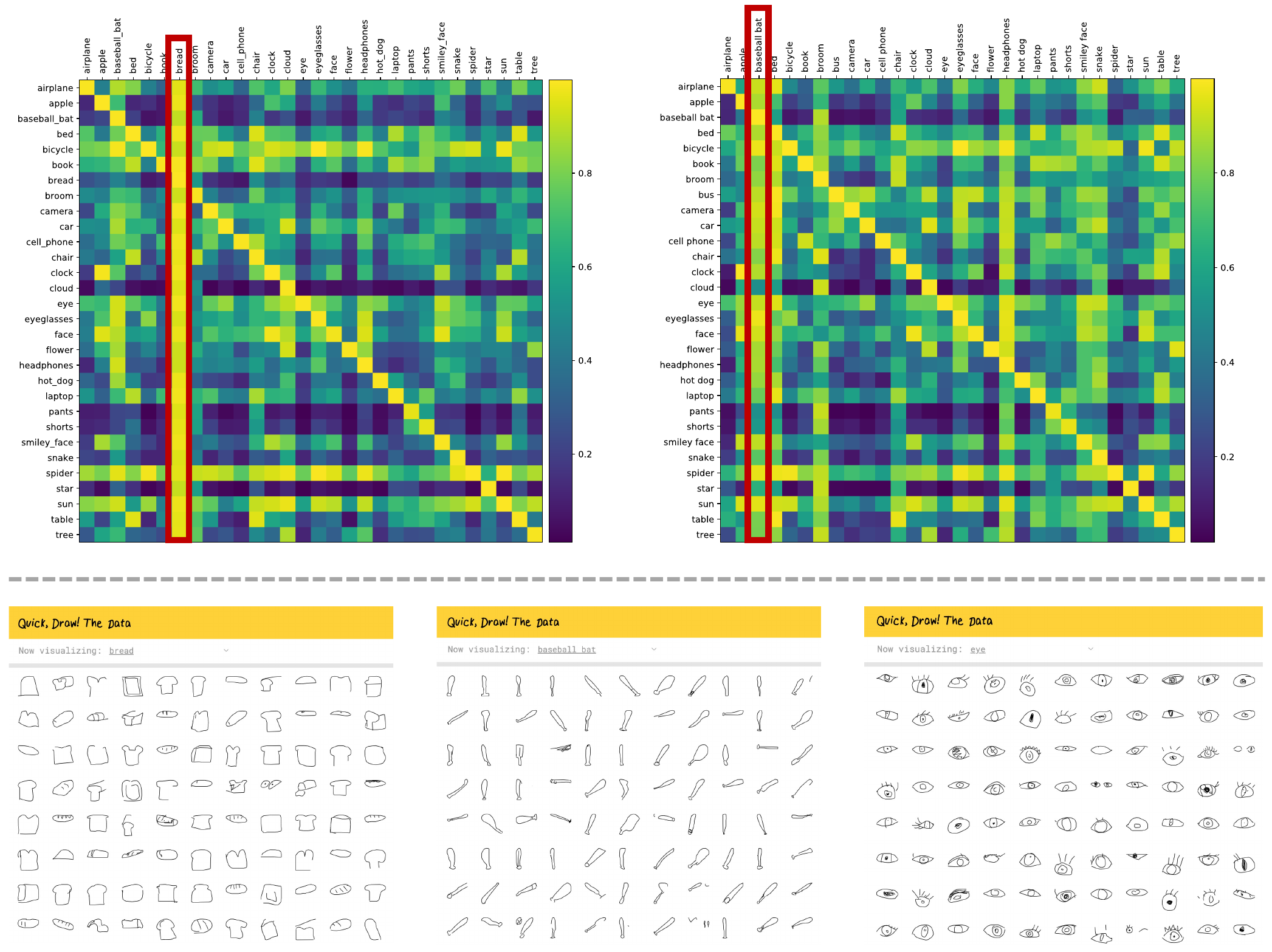}
    \caption{
    \textbf{Transfer maps and examples.} Top: we apply SLI on transfer tasks between every two categories out of a total of 30 and observe all sketch samples regardless of the origin can be transferred to \texttt{[bread]} (left). To confirm, we exclude \texttt{[bread]} and replace it with a new category \texttt{[bus]} and this time all sketches transfer to \texttt{[baseball\_bat]}. Bottom: we showcase the screenshots of three QuickDraw categories, \texttt{[bread]}, \texttt{[baseball\_bat]}, \texttt{[eye]}, which yields an explanation to the said phenomenon.
    }
    \vspace{3em}
    \label{fig:transfer_map}
\end{figure}

To explore the link between the categories, we build the transfer map based on SLI. At the top of \cref{fig:transfer_map}, the transfer map donates the average confidence between the every two categories of a total of 30 in the transfer SLI task. We find some interesting phenomena: (i) The diagonal line is highlighted, indicating that it is the recovery task rather than the transfer one, and therefore has the highest average confidence. (ii) Sketches in the categories with a high number of strokes are easily transferred into other categories and vice versa (see \texttt{[spider]} and \texttt{[star]}). (iii) When we set some categories as the target categories for the transfer task in SLI, we get very high average confidence (see \texttt{[bread]} and \texttt{[baseball\_bat]}).

\section{Effect of different location initialisation on explanations}

\begin{figure} [H]
    \centering
    \vspace{3em}
    \includegraphics[width=\linewidth]{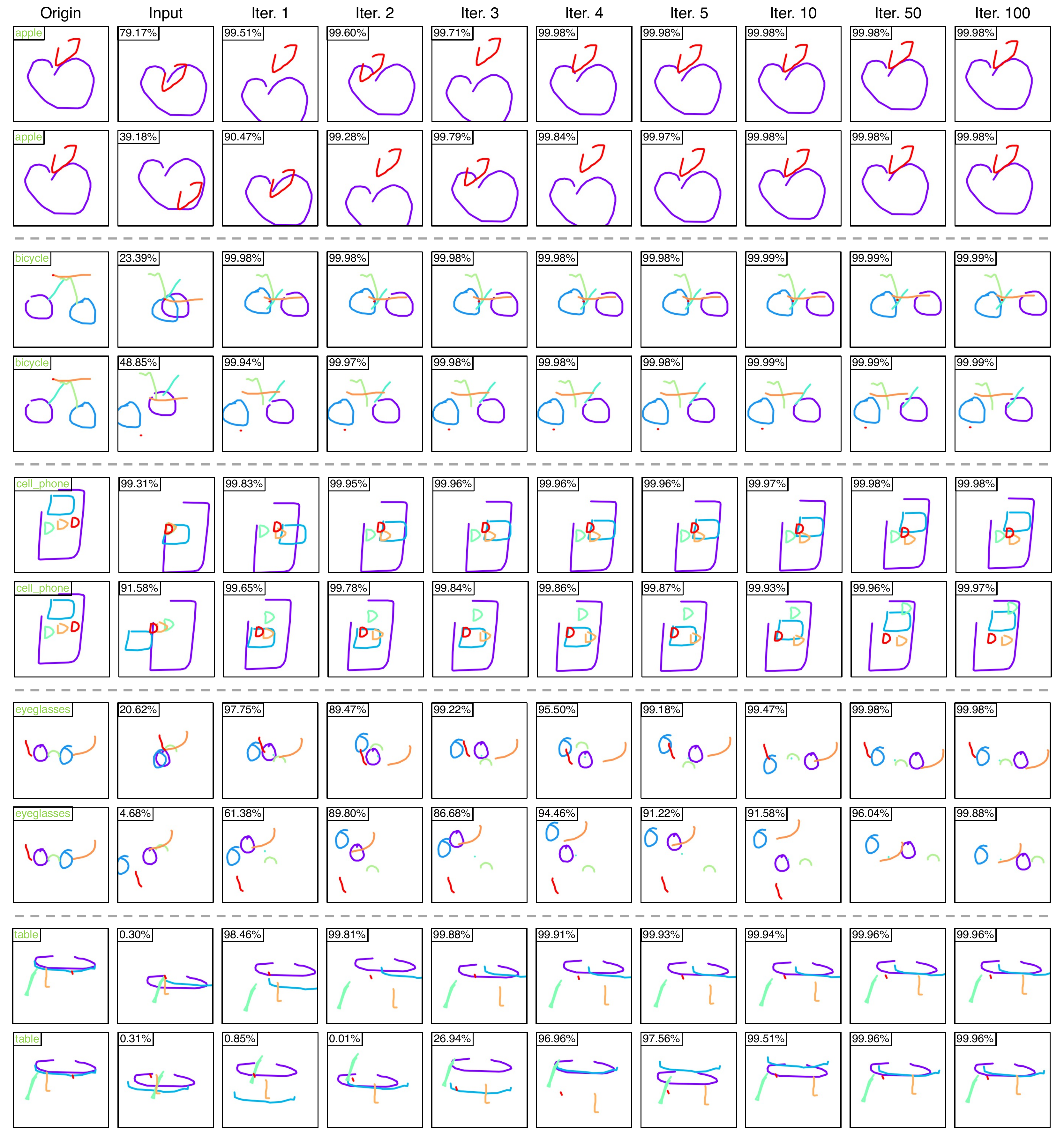}
    \caption{
    Here we show the visualisations of the 100 optimisations of the model inversion. For each sketch, the first row represents the optimisation process where the locations are initialised at the centre of the canvas. While the second row represents the optimisation process with random initialisation of locations. The number in the top-left corner indicates the confidence that the model predicts the current sketch belongs to the original label.
    }
    \label{fig:recovery_comparison}
\end{figure}

To explore the effect of different initialization locations on explanations, we design the following comparative experiments: (i) Reset location information of all strokes to 0, \ie, all starting points are placed in the centre of the canvas. (ii) Randomly place strokes according to a normal distribution. In \cref{fig:recovery_comparison}, we find that different initialization locations will affect the entire recovery process and generate different sketches. However, since the shapes of strokes do not change, the final structures of sketches are similar.

\section{Additional visualisations}

\vspace{-1.5em}

\begin{figure} [H]
    \centering
    \begin{subfigure}[h]{0.9\textwidth}
        \includegraphics[width=\textwidth]{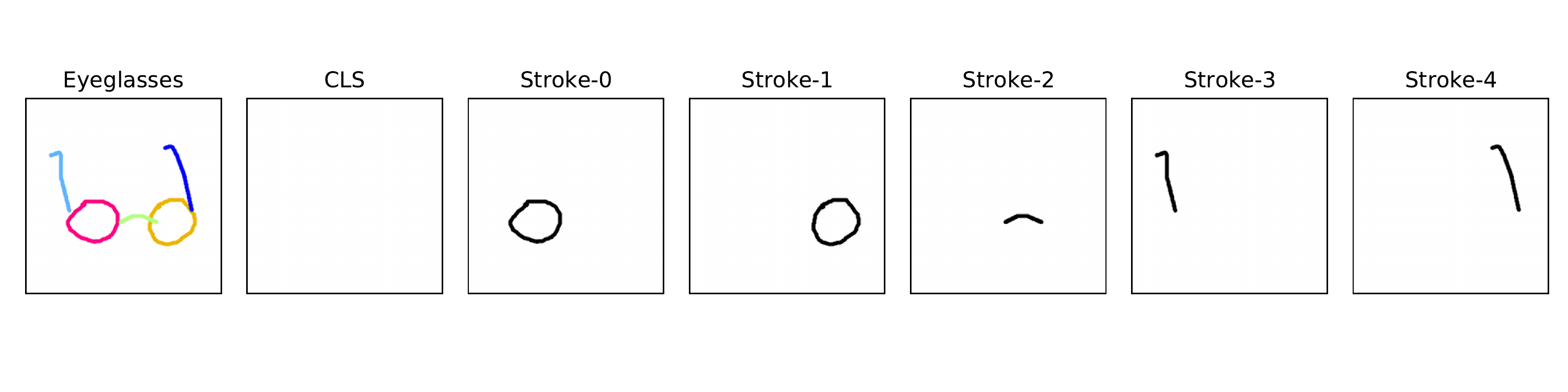}
        \vspace{-3.5em}
        \caption{A sketch sample of eyeglasses and the 6 tokens it contains (1 CLS token and 5 stroke tokens).}
    \end{subfigure}%

    \begin{subfigure}[h]{0.3\textwidth}
        \includegraphics[width=\textwidth]{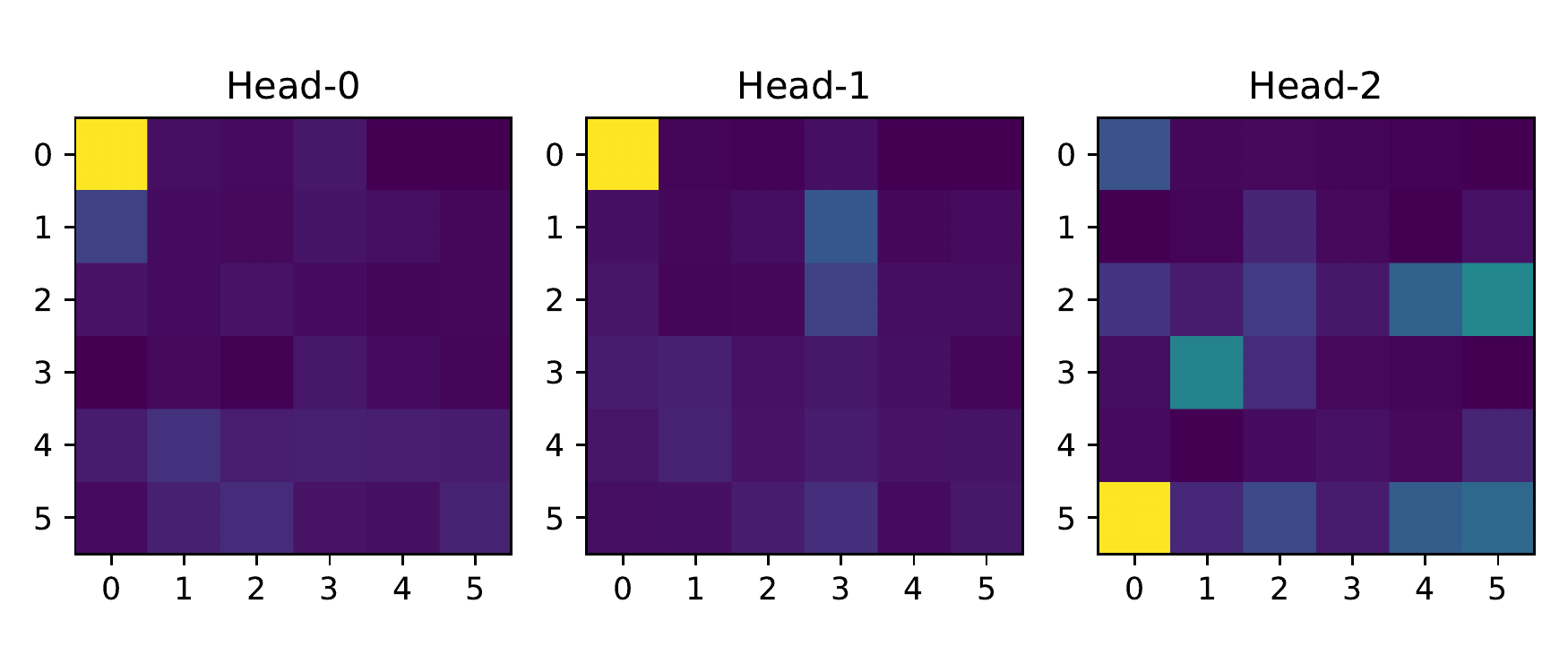}
        \caption{Layer-0}
    \end{subfigure}%
    \quad
    \begin{subfigure}[h]{0.3\textwidth}
        \includegraphics[width=\textwidth]{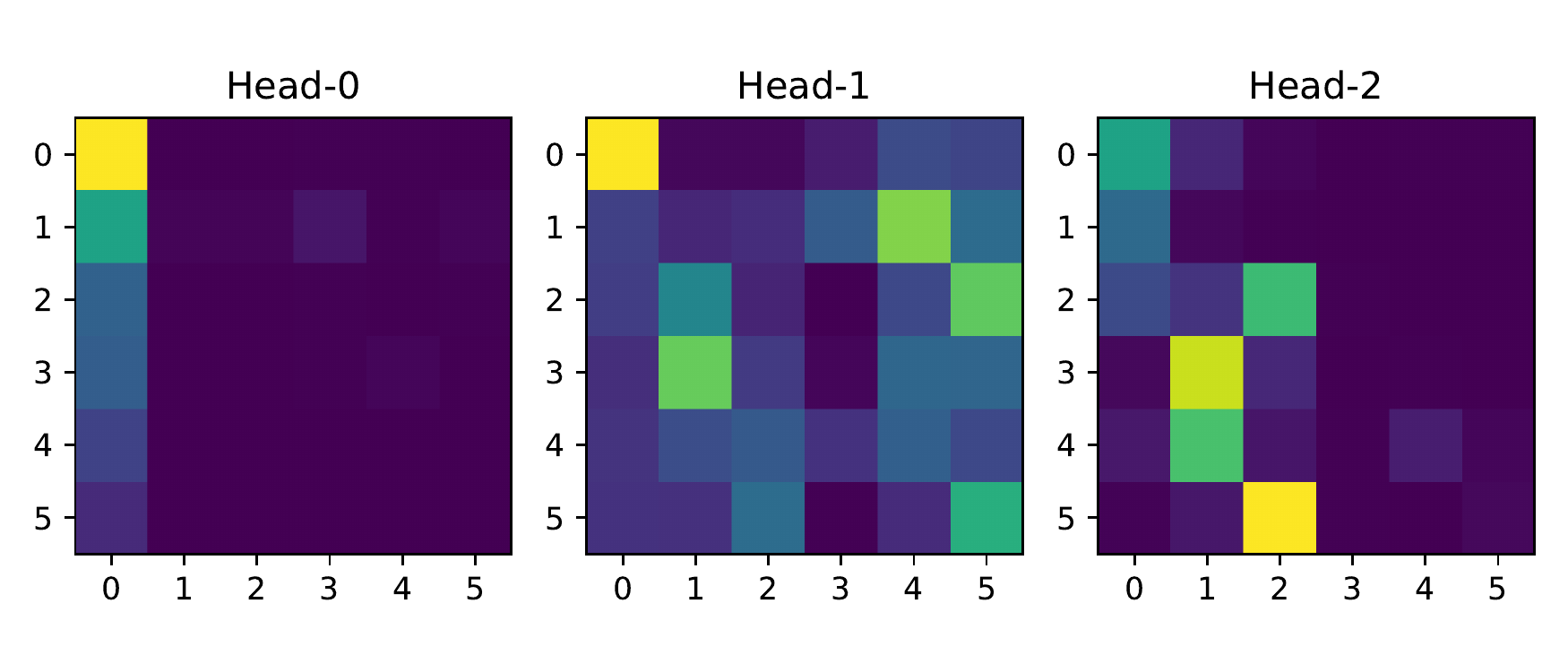}
        \caption{Layer-1}
    \end{subfigure}
    \quad
    \begin{subfigure}[h]{0.3\textwidth}
        \includegraphics[width=\textwidth]{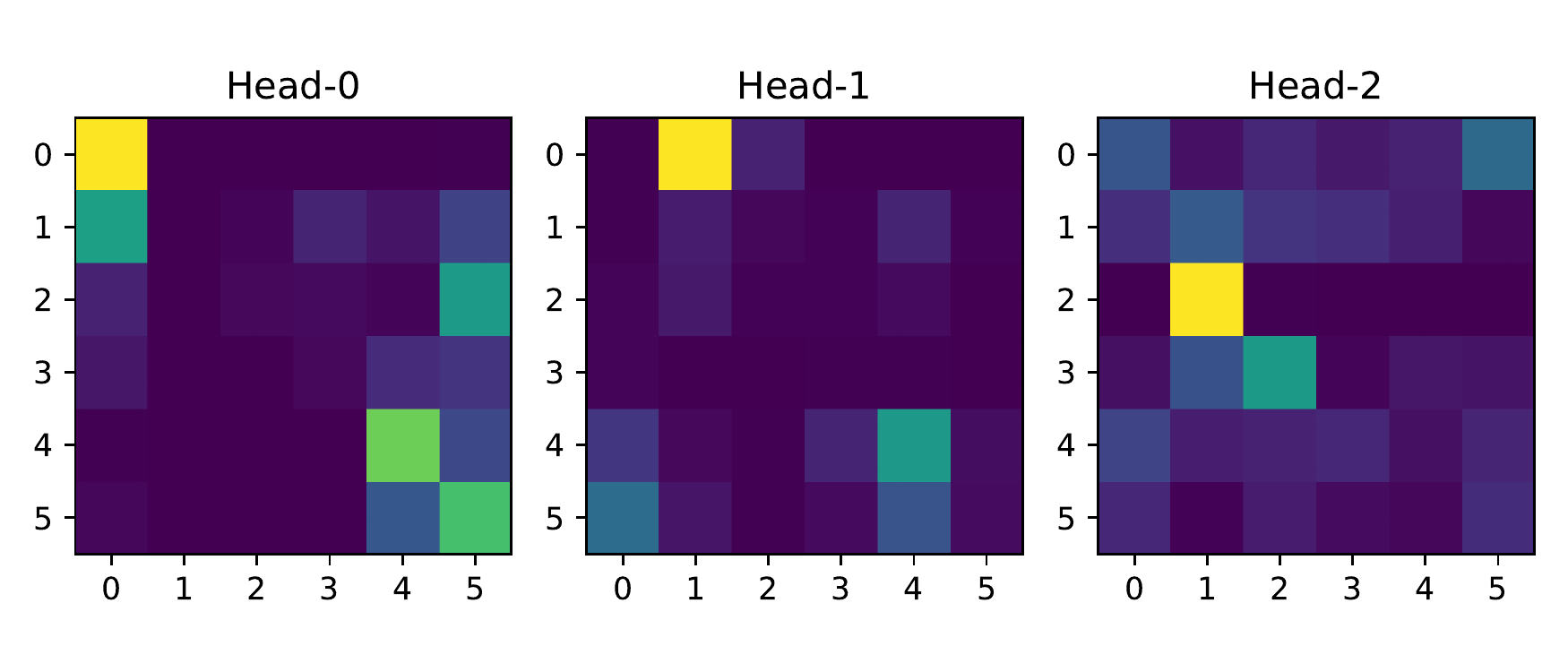}
        \caption{Layer-2}
    \end{subfigure}

    \begin{subfigure}[h]{0.3\textwidth}
        \includegraphics[width=\textwidth]{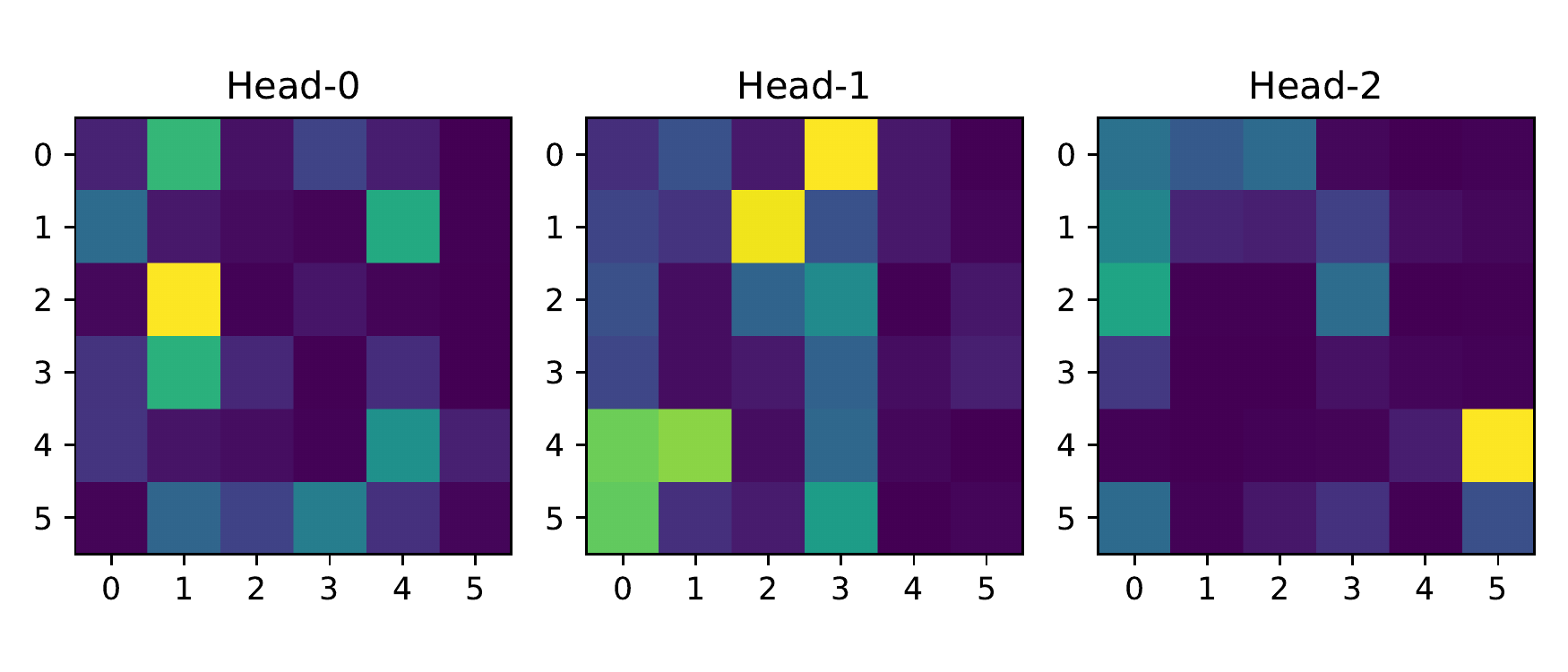}
        \caption{Layer-3}
    \end{subfigure}%
    \quad
    \begin{subfigure}[h]{0.3\textwidth}
        \includegraphics[width=\textwidth]{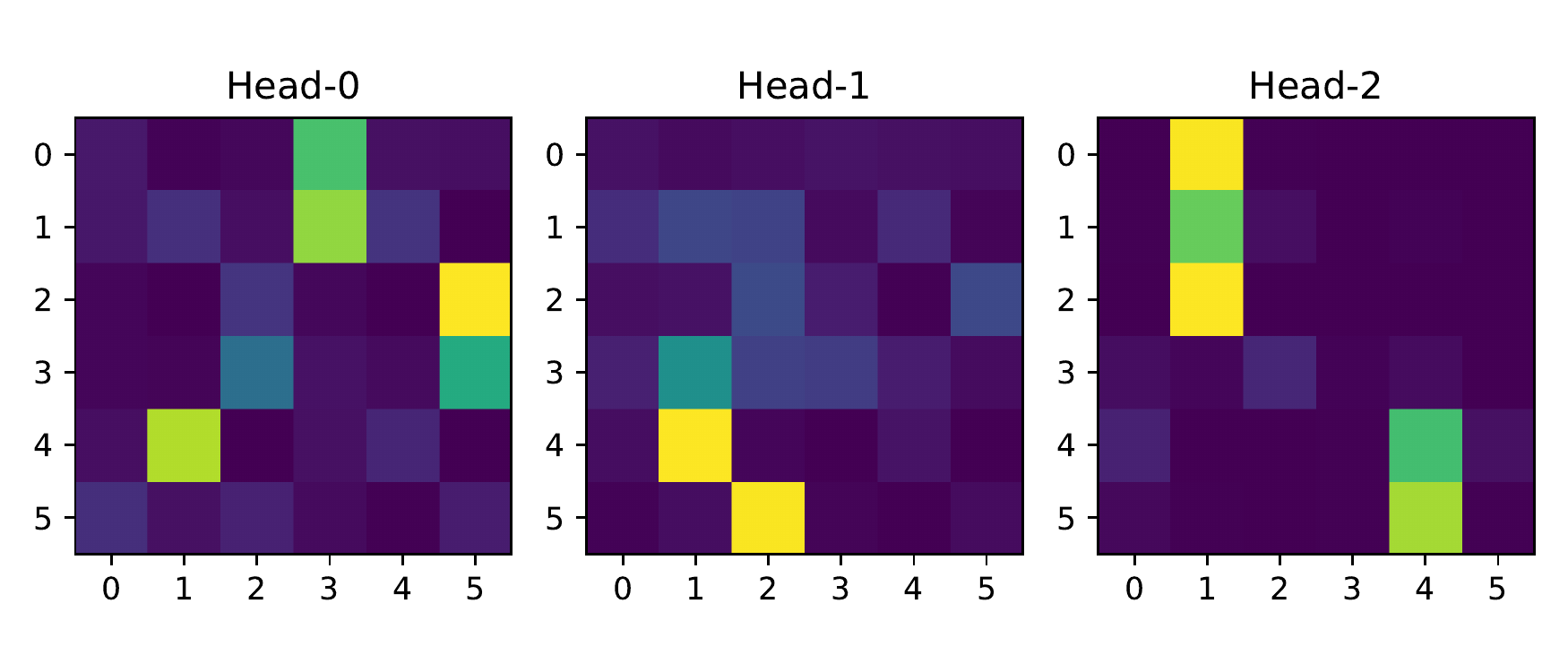}
        \caption{Layer-4}
    \end{subfigure}
    \quad
    \begin{subfigure}[h]{0.3\textwidth}
        \includegraphics[width=\textwidth]{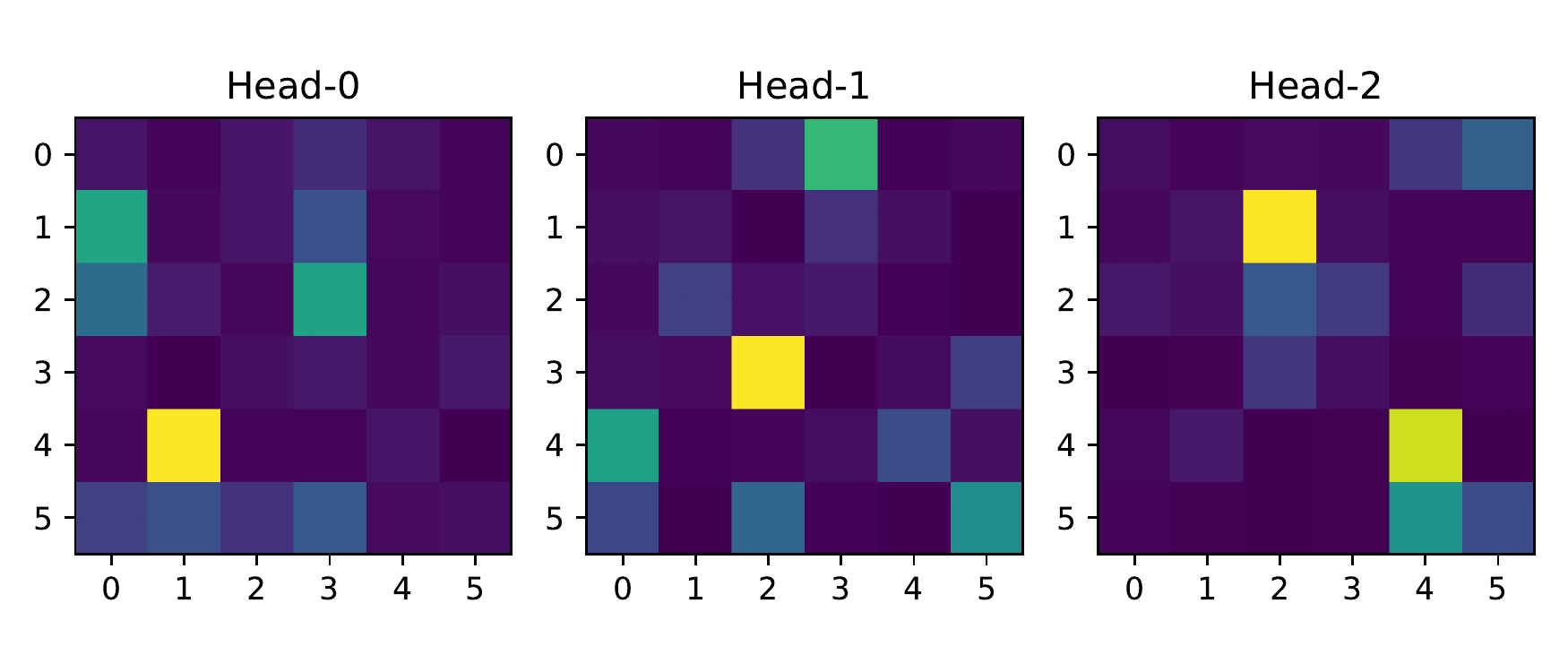}
        \caption{Layer-5}
    \end{subfigure}

    \begin{subfigure}[h]{0.3\textwidth}
        \includegraphics[width=\textwidth]{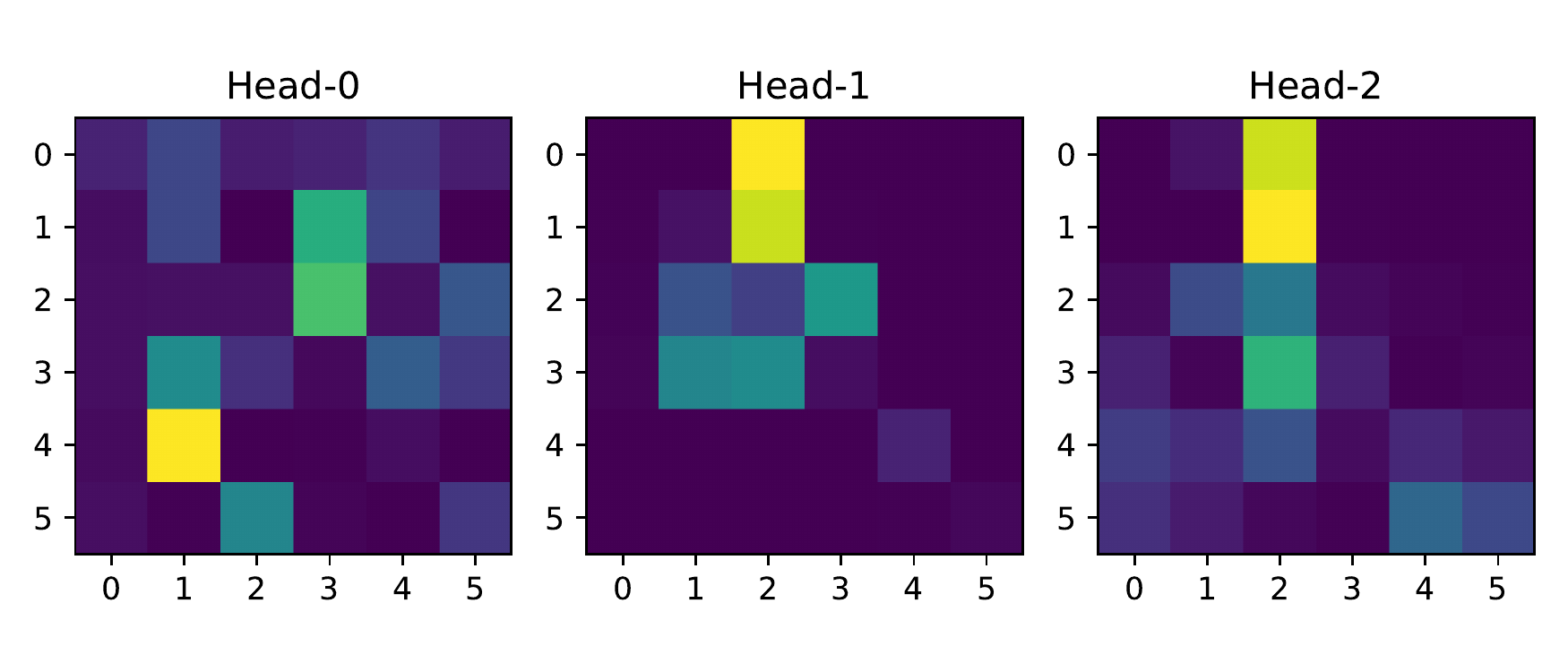}
        \caption{Layer-6}
    \end{subfigure}%
    \quad
    \begin{subfigure}[h]{0.3\textwidth}
        \includegraphics[width=\textwidth]{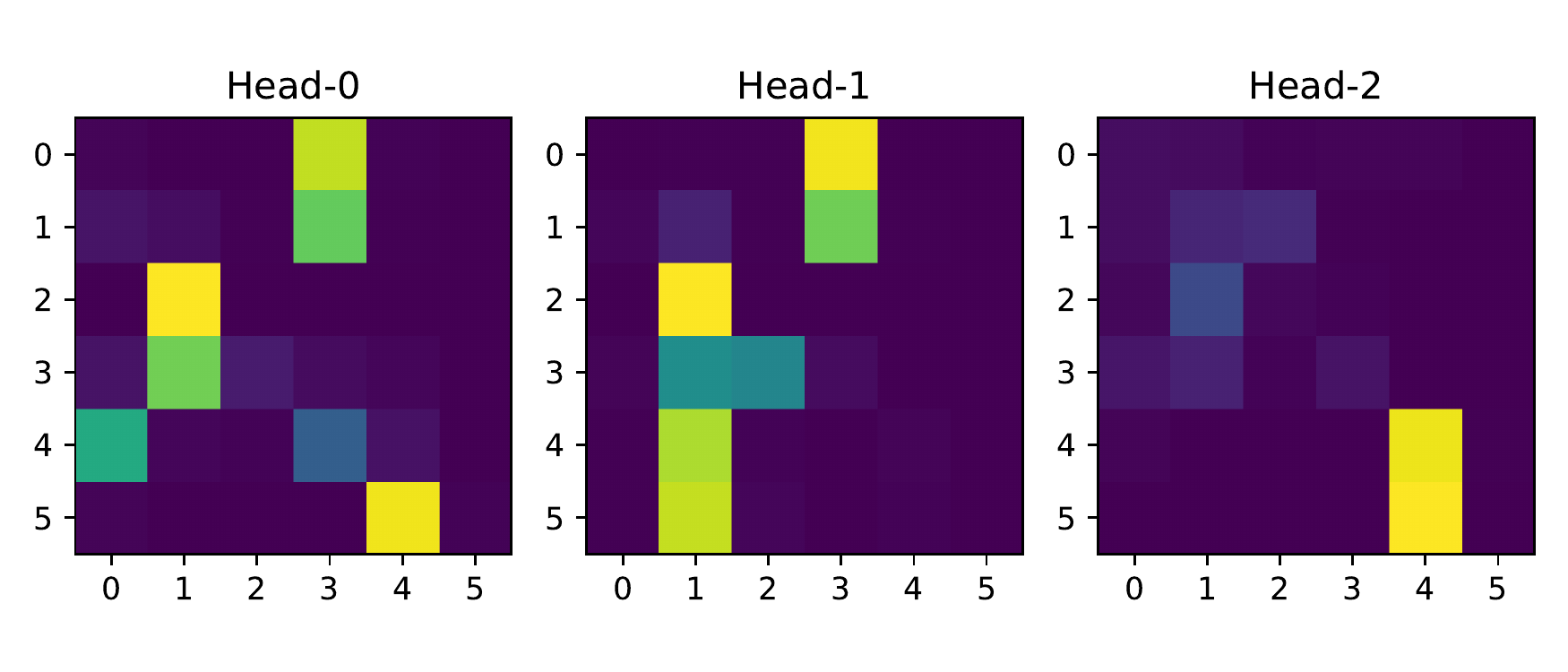}
        \caption{Layer-7}
    \end{subfigure}
    \quad
    \begin{subfigure}[h]{0.3\textwidth}
        \includegraphics[width=\textwidth]{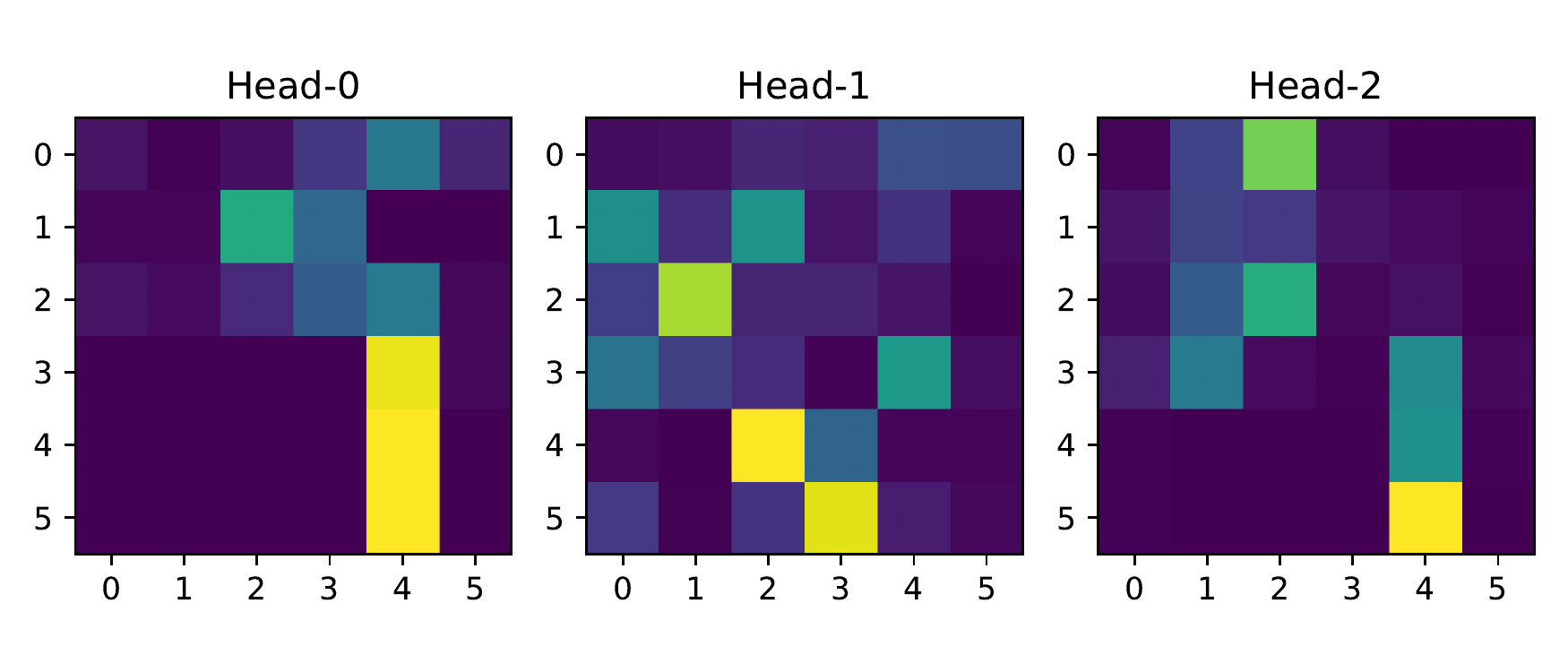}
        \caption{Layer-8}
    \end{subfigure}

    \begin{subfigure}[h]{0.3\textwidth}
        \includegraphics[width=\textwidth]{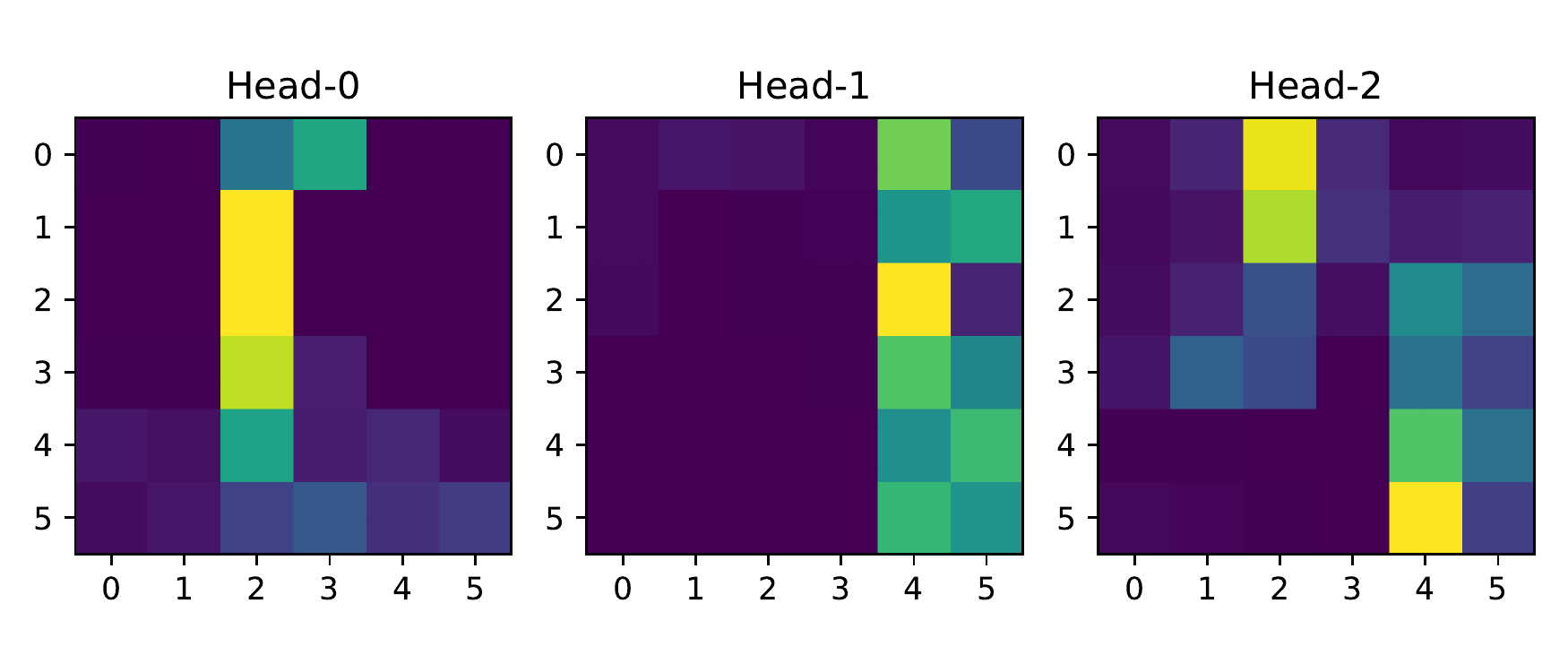}
        \caption{Layer-9}
    \end{subfigure}%
    \quad
    \begin{subfigure}[h]{0.3\textwidth}
        \includegraphics[width=\textwidth]{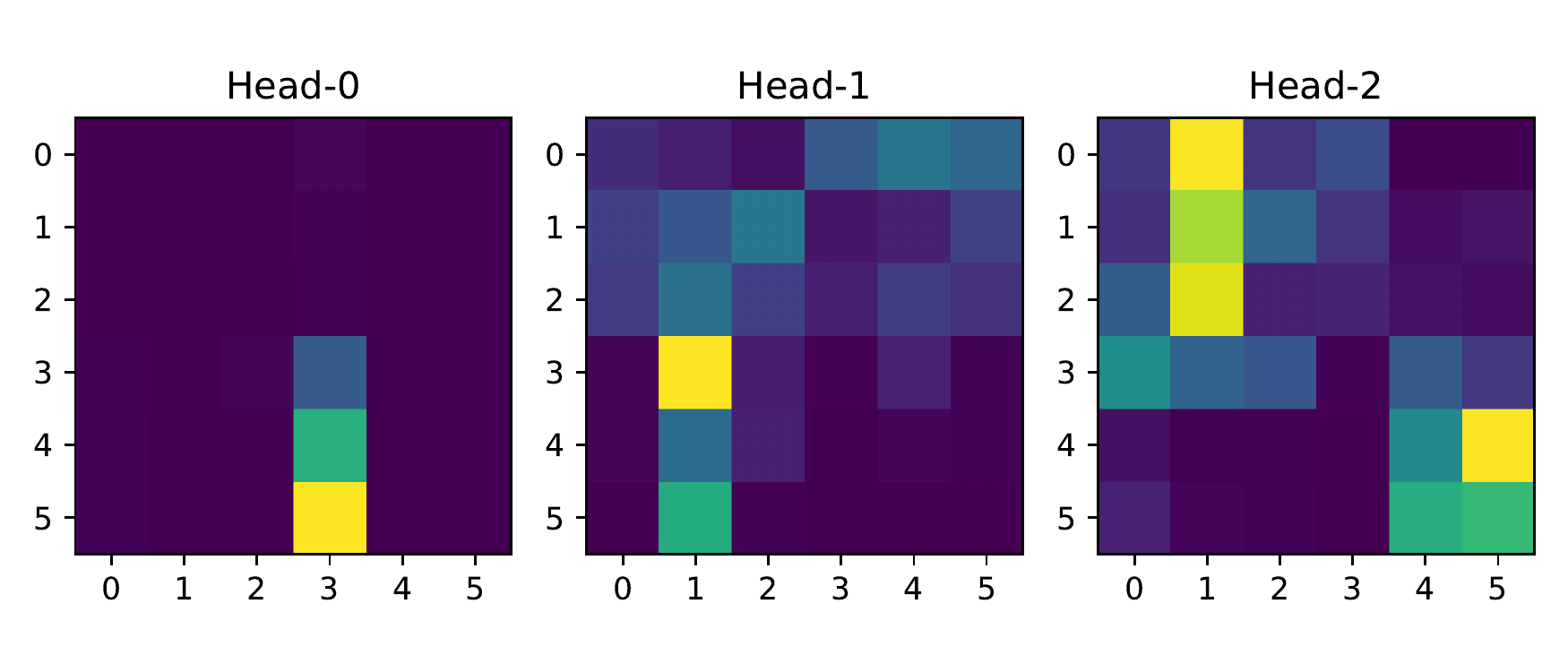}
        \caption{Layer-10}
    \end{subfigure}
    \quad
    \begin{subfigure}[h]{0.3\textwidth}
        \includegraphics[width=\textwidth]{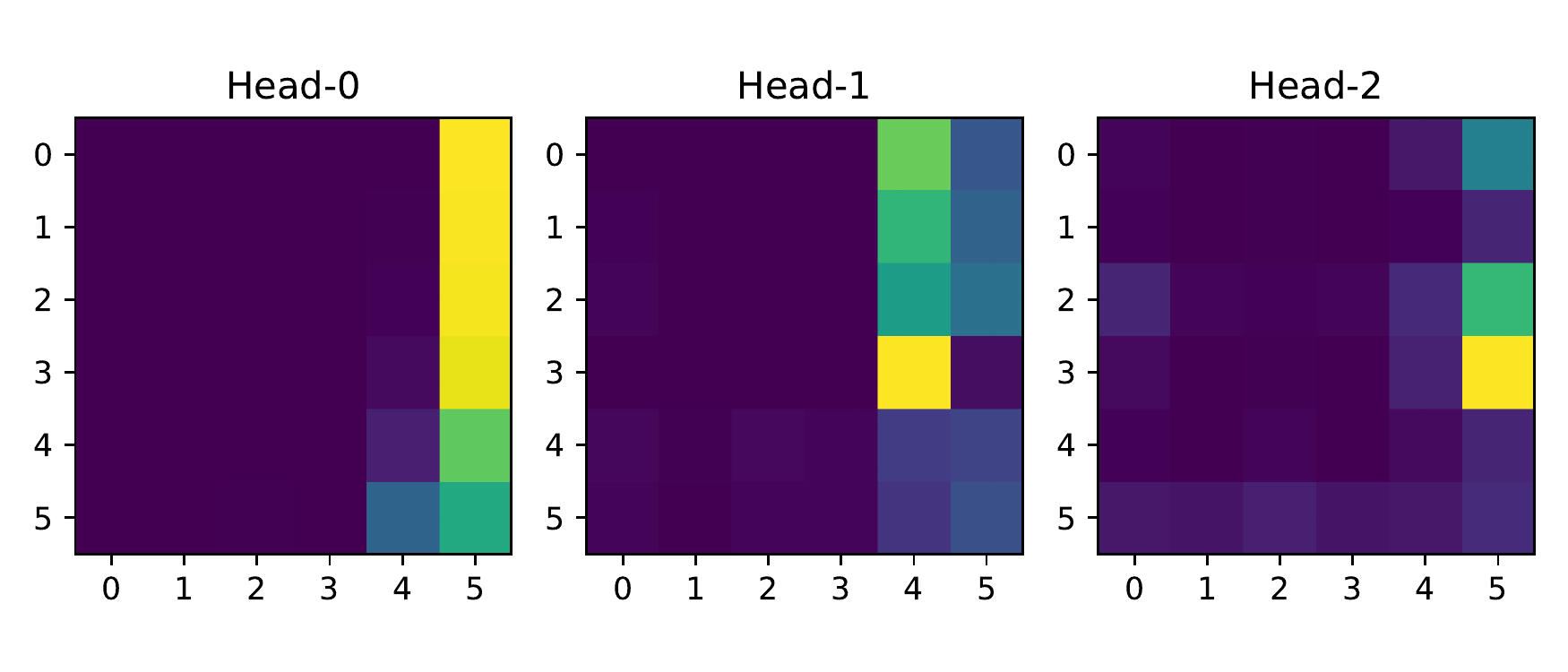}
        \caption{Layer-11}
    \end{subfigure}

    \caption{Visualisation of attentions in different layers of Transformer Encoder between strokes.}\label{fig:attention}
\end{figure}

\end{document}